\documentclass[11pt,letterpaper]{article}
\usepackage[margin=1in]{geometry}
\usepackage{times,natbib}
\setcitestyle{authoryear,round,citesep={;},aysep={,},yysep={;}}
\usepackage{amsmath,amssymb,graphicx,booktabs,multirow,array}
\usepackage{float,caption,microtype,wrapfig,flafter}
\usepackage{algorithm,algpseudocode,xcolor,tikz}
\usepackage{hyperref,url}
\definecolor{stageA}{HTML}{287A77}
\definecolor{stageB}{HTML}{326BA5}
\definecolor{stageC}{HTML}{8756A5}
\definecolor{stageD}{HTML}{596575}

\newcommand{\algstage}[2]{\Statex\hspace{-\algorithmicindent}\colorbox{stage#1!10}{\parbox{\dimexpr\linewidth-2\fboxsep\relax}{\textcolor{stage#1}{\textbf{Stage #1}}\quad\textbf{#2}}}}
\newcommand{\alghelper}[1]{\Statex\hspace{-\algorithmicindent}\colorbox{stageD!8}{\parbox{\dimexpr\linewidth-2\fboxsep\relax}{\textcolor{stageD}{\textbf{Shared episode}}\quad\textbf{#1}}}}
\graphicspath{{figures/}}

\newcommand{\finalpr}{63.75}
\newcommand{\finalauc}{71.13}
\newcommand{\finalf}{60.56}
\newcommand{\finalcells}{30}
\newcommand{\finaltasks}{12}

\hypersetup{hidelinks,pdfauthor={Tian Zhou, Bingqing Peng, Linxiao Yang, Wenwei Wang, Mengni Ye, Beverly Jin, Zuyi Zhu, Jinjie Gu, Liang Sun},pdftitle={Transferable Evidence Reconstruction for Longitudinal Glucose Representations}}
\title{Transferable Evidence Reconstruction for\\Longitudinal Glucose Representations}
\author{Tian Zhou$^1$, Bingqing Peng$^1$, Linxiao Yang$^1$, Wenwei Wang$^1$, Mengni Ye$^2$\\
Beverly Jin$^2$, Zuyi Zhu$^1$, Jinjie Gu$^1$, Liang Sun$^1$\\[0.5em]
\small $^1$Ant Group \qquad $^2$Independent Researcher}

\date{}
\begin{document}
\maketitle
\begin{abstract}
\looseness=-1
Long physiological recordings contain many routine measurements, while predictive information often lies in rare events, sustained burden, and recurring patterns. These properties can be computed as label-free evidence, but directly using them as features leaves limited labeled data to separate reproducible associations from sample-specific ones. Learning to reconstruct evidence can exploit unlabeled recordings, yet joint reconstruction does not explicitly require the decoding rule to transfer across individuals. We introduce \emph{transferable evidence reconstruction} (TER): a Ridge regressor fits evidence from representations in one group and predicts it in an identity-disjoint group without refitting. The transfer error trains the encoder through the differentiable fit. For continuous glucose monitoring (CGM), clock-aware encoding preserves the multi-day content and timing needed for evidence recovery. Matched interventions connect the gains to reduced fitting-group sensitivity, with structured targets improving on raw recovery. Across ten leading CGM and time-series baselines, TER sets a new best metric on \textbf{\finaltasks{}/14 phenotype tasks} and exceeds the strongest prior overall PR-AUC/ROC-AUC/Macro-F1 by \textbf{4.95/4.43/0.66 percentage points}; the PR-AUC and ROC-AUC gains are \textbf{2.6$\times$ and 2.2$\times$} the respective gaps between the two strongest baselines. Meal-response and future-CGM studies further demonstrate predictive utility. TER thus uses meaningful signal properties to supervise not only what a representation preserves, but how reliably it can be read across individuals.
\end{abstract}

\section{Introduction}

A week of glucose measurements can reveal something that a single day cannot: whether an elevation is an isolated episode or a recurring burden. Yet a longer recording also brings thousands of routine measurements. The useful distinction may lie not in any one value, but in how often an event returns, how long recovery takes, and when it happens. More observations therefore create an opportunity, not a guarantee, for better representations. How should self-supervision turn this history into information that remains useful in new individuals?

Standard self-supervised objectives answer related questions: contrastive learning asks which views should agree, and masked autoencoding asks which missing measurements can be recovered \citep{chen2020simclr,he2022mae}. Neither explicitly asks whether occasional elevations can be distinguished from recurring ones. Physiological recordings offer a useful opportunity here: burden, extremes, and recurrence can often be measured without knowing a person's phenotype \citep{battelino2019tir,hall2018glucotypes}. We call these signal-derived properties \emph{structured evidence}.

This suggests an appealing shortcut: compute the evidence and use it directly for prediction. But specifying a meaningful measurement does not determine how to represent its variation. A small labeled cohort must still separate reproducible differences from sample-specific associations. We instead use evidence to supervise learning across many unlabeled recordings, leaving the encoder to organize the information before a downstream task is fitted.

Changing the reconstruction target is only part of the answer. A jointly trained encoder and decoder can learn to work well together, yet downstream use discards that decoder and fits a new predictor from a small group of people. Reconstruction accuracy alone does not test this change of fitting population. Our central question is therefore: \emph{can a simple evidence mapping learned from one group also work on another, without being fitted again?}

\begin{figure}[t]\centering
\includegraphics[width=\linewidth]{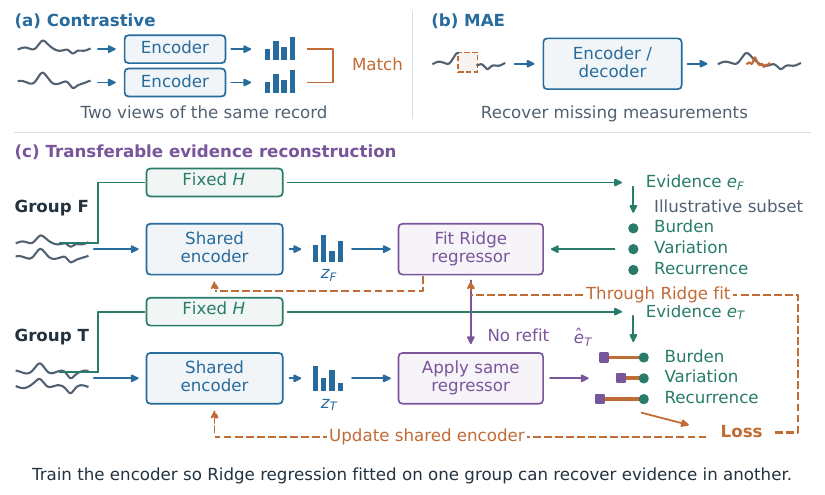}
\caption{\textbf{TER combines recovery with an explicit transfer test.} Contrastive learning specifies views that should agree; MAE specifies measurements that should be recoverable. TER recovers structured evidence, but scores it through a Ridge regressor fitted on other identities. In (c), $r_F^\star$ never fits or recalibrates on $E_T$; gradients pass through both groups and the episodic fit, and only the encoder is retained.}
\label{fig:ssl-comparison}
\end{figure}

\emph{Transferable evidence reconstruction} (TER) turns this question into a training objective (Figure~\ref{fig:ssl-comparison}). At each update, a Ridge regressor learns to read evidence from one group of recordings and is applied unchanged to an identity-disjoint group. Its transfer error updates the encoder through the differentiable fit. The representation is thus trained not just to support recovery, but to support a rule learned elsewhere. Our interventions connect this requirement to reduced fitting-group sensitivity and more consistent class differences across groups.

For CGM, this brings us back to the original week-long recording: the encoder must retain the relations the objective asks it to recover. Pooling daily embeddings loses their order; retaining order with a sequence model does not by itself reward burden or recurrence. We bind observed glucose content to physical time before combining days and train the resulting representation with TER. Temporal organization preserves the relevant history; cross-group recovery teaches the encoder to make its evidence transferable.

Our contributions are:
\begin{enumerate}
\item \textbf{Evidence recovery with an explicit transfer requirement.} TER uses signal-derived properties as supervision rather than fixed downstream features, and learns their recovery through a simple regressor fitted on other individuals. Matched interventions distinguish the benefit of cross-group scoring from reconstruction accuracy alone.
\item \textbf{Temporal organization for longitudinal evidence.} Our CGM encoder retains observed daily content and its clock relations across days. Same-history and objective controls explain why this combination improves on simply pooling daily representations or extending the input.
\item \textbf{Comprehensive representation evaluation.} We cover GlucoFM's complete public phenotype benchmark: four cohorts, seven phenotype families, and 14 cohort--task pairs. Against ten CGM and time-series baselines, TER leads \textbf{\finaltasks{}/14 tasks} and exceeds GlucoFM by \textbf{4.95/4.43/0.66 points} in overall PR-AUC/ROC-AUC/Macro-F1. Private-cohort meal responses and public-trial future outcomes extend evaluation beyond this benchmark.
\end{enumerate}

\section{Related work}

\paragraph{Self-supervised targets and episodic learning.}
Contrastive learning matches constructed views, whereas masked autoencoding reconstructs omitted observations \citep{chen2020simclr,he2022mae}. Video SSL has predicted signal-derived statistics \citep{wang2019videostatistics}; PULSE cross-reconstructs augmented physiological windows to retain shared dynamics \citep{chen2026pulse}. Differentiable Ridge regression and inner--outer optimization appear in labeled meta-learning \citep{bertinetto2019r2d2,lee2019metaoptnet,franceschi2018bilevel}; CACTUs and STUNT construct unlabeled pseudo-tasks \citep{hsu2019cactus,nam2023stunt}. TER brings these ideas together around continuous evidence from physiological recordings. Its learning task is to recover this evidence across groups using a newly fitted regressor, connecting signal-derived supervision with the transfer requirement of episodic learning.

\paragraph{Longitudinal health and CGM representations.}
Irregular-series models represent observation indices, missingness, or continuous-time dynamics \citep{li2020raindrop,moe2021larnn}. CGMformer reconstructs masked daily tokens, GluFormer predicts glucose autoregressively, CGM-JEPA predicts masked latents, and GlucoFM separates slow state from transient events \citep{lu2025cgmformer,lutsker2026gluformer,muhammad2026cgmjepa,li2026glucofm}. Our focus is how observation-supported daily information becomes a useful multi-day representation. The encoder forms glucose level/change--clock relations before aggregation, and TER trains the resulting representation for evidence recovery across recording groups.

\section{Transferable evidence reconstruction}
\label{sec:formulation}

We first specify which signal properties to preserve, then require their recovery across individuals. Section~\ref{sec:longitudinal-encoder} builds the multi-day encoder that retains the necessary content and timing.

\subsection{What to recover: signal-derived evidence}

To give long-term properties an explicit role in learning, we define a fixed evidence function $H$. Given a signal and its observation context $x$, the encoder produces $z=f_\theta(x)$ and $H$ produces $e=H(x)$. The target describes properties of the recording, not downstream phenotype labels. It supervises the encoder rather than replacing its output with hand-computed features.

A concrete example is repeated high-glucose burden. For day $d$, let
\begin{equation}
q_d=\frac{\sum_i m_{d,i}\,\mathbf1[g_{d,i}>180]}
{\max(1,\sum_i m_{d,i})},
\label{eq:evidence-example}
\end{equation}
where $g$ is glucose in mg/dL and $m$ indicates observation. Across days, the support-weighted mean and spread of $q_d$, together with recurrence of high-glucose days 24 hours apart, describe typical burden, variability, and persistence, giving brief events and sustained patterns explicit weight.

The target contains 64 current-day, 256 longitudinal, and 16 low/high-event summaries (Appendix~\ref{sec:target-details}). Why not use these values directly as features? Their variation includes both reproducible differences and associations specific to the fitting sample. Using evidence as a target lets unlabeled data shape the representation before downstream fitting. Section~\ref{sec:learning-audit} tests this distinction by changing how the same summaries are represented, without adding information. Evidence specifies what learning should preserve, not the final feature space.

\subsection{How to learn it: fit on one group, recover in another}

With the target defined, the simplest approach jointly trains an encoder and decoder:
\begin{equation}
J_{\mathrm{joint}}(\theta,w)=
\mathbb E_{x}\!\left[\ell\!\left(w(f_\theta(x)),H(x)\right)\right].
\label{eq:joint-decoder-objective}
\end{equation}
Joint reconstruction optimizes one encoder--decoder pair. It does not explicitly test whether re-estimating the decoder on different samples yields consistent predictions. TER introduces this test: fit a regularized linear readout on one group and score it on other individuals. A close fit with an unreliable transferred rule receives a high loss. Section~\ref{sec:why} tests this sensitivity and the effect of cross-group scoring.

At every update, we sample fit and transfer groups $F$ and $T$ with disjoint recorded identities. Let $Z_F=f_\theta(X_F)$ and $E_F=H(X_F)$, and define $Z_T,E_T$ analogously. A Ridge regressor is fitted on $F$ and then applied unchanged to $T$:
\begin{align}
r_F^\star(\theta)
&=\operatorname{Ridge}\bigl(Z_F,E_F\bigr),\nonumber\\
J(\theta)
&=\mathbb E_{F,T}\!
 \left[\ell\!\left(r_F^\star(\theta)(Z_T),E_T\right)\right],
\qquad g(F)\cap g(T)=\varnothing .
\label{eq:generic-objective}
\end{align}
Here $g$ denotes recorded identity and $\ell$ is weighted squared error. The regressor and episode-specific scaling are estimated on $F$; the targets $E_T$ supply the transfer loss. Both groups come from the unlabeled pretraining fitting pool. Because the Ridge solve is differentiable, the loss updates $Z_T$ directly and $Z_F$ through $r_F^\star(\theta)$. The regressor is then discarded. Appendix~\ref{sec:ridge} specifies normalization and the exact solve; Appendix~\ref{sec:identity-scope} describes recorded identities across data sources.

\begin{algorithm}[H]
\caption{One update of transferable evidence reconstruction}
\label{alg:ter-episode}
\small
\begin{algorithmic}[1]
\Require Encoder $f_\theta$, unlabeled fitting pool, fixed evidence function $H$.
\State \textbf{Sample:} draw groups $F,T$ with disjoint recorded identities.
\State \textbf{Encode:} compute $Z_F=f_\theta(X_F)$ and $Z_T=f_\theta(X_T)$.
\State \textbf{Construct targets:} compute $E_F=H(X_F)$ and $E_T=H(X_T)$.
\State \textbf{Fit:} obtain an episodic differentiable Ridge regressor $r_F^\star$ from $(Z_F,E_F)$ only.
\State \textbf{Transfer:} predict $\widehat E_T=r_F^\star(Z_T)$ without using $E_T$ to fit or recalibrate.
\State \textbf{Learn:} update the active encoder parameters using
$\nabla_\theta\ell(\widehat E_T,E_T)$, including the gradient through the fit.
\State Discard this regressor before the next update.
\end{algorithmic}
\end{algorithm}

\section{Applying TER to multi-day glucose recordings}
\label{sec:longitudinal-encoder}

\begin{figure}[H]\centering
\includegraphics[width=\linewidth]{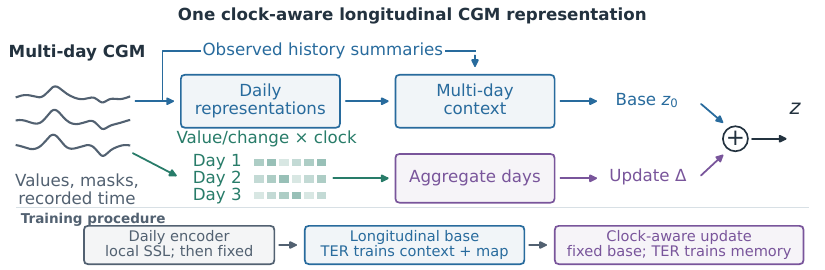}
\caption{\textbf{A CGM encoder for transferable evidence reconstruction.} Daily representations and supported history summaries form $z_0$. Clock-aware memory binds glucose level/change to physical time before pooling days and supplies an additive update $\Delta$. The final representation preserves the content and temporal relationships needed to recover evidence. Local daily learning precedes TER training of the longitudinal base and then the clock-aware update.}
\label{fig:architecture}
\end{figure}

We now specify $f_\theta$ for multi-day CGM. More input provides access to history, but does not specify which relations should survive compression. Mean--max pooling removes day order; a BiLSTM can retain it, but its objective determines what is rewarded. We couple value--time organization with evidence-recovery training to preserve properties spanning multiple days (Figure~\ref{fig:architecture}). Section~\ref{sec:history-use} tests these alternatives and changes the objective within the same BiLSTM.

\subsection{Preserve what each day actually observes}

Recurrence can only be interpreted relative to what was observed each day. The daily encoder therefore reads glucose with observation masks and sampling information, preserving glucose level and local changes without treating missing values as observations. It is learned without phenotype labels before longitudinal training; Appendix~\ref{sec:daily-implementation} gives its sampling-consistency and local-dynamics objectives.

TER next trains a context encoder to combine daily representations with observed-history summaries and elapsed day age, producing $z_0(x)$. This captures multi-day content. However, summarizing days can obscure when particular values occurred, motivating an additional path that explicitly retains their clock positions.

\subsection{Bind content to clock position before combining days}

Mean--max pooling is invariant to the order of daily embeddings. A recurrent encoder can retain order, but clock-aware memory additionally makes the relation between glucose content and physical time explicit before pooling. Sine and cosine weights summarize where that content falls within a day, distinguishing, for example, morning from evening elevations with the same glucose values.

Let $b_{d,i}$ contain fixed bounded transforms of glucose level and adjacent glucose change on day $d$, and let $\theta_{d,i}$ denote recorded time of day. We form clock summaries
\begin{equation}
s_{d,k}=
\frac{\sum_i m_{d,i}
 [b_{d,i}\cos(k\theta_{d,i}),\;b_{d,i}\sin(k\theta_{d,i})]}
{\max(1,\sum_i m_{d,i})}.
\label{eq:binding}
\end{equation}
The mask prevents changes from crossing missing readings, and multiple harmonics retain within-day arrangement at several scales. Attention then combines valid day--harmonic tokens with elapsed-age information. Unlike appending a timestamp after pooling, Equation~\ref{eq:binding} forms the level/change--clock relation before any day is aggregated. Appendix~\ref{sec:memory-implementation} gives the fixed transforms and attention operations.

\subsection{Combine daily content and temporal information}

The clock-aware path complements the content already represented by $z_0$, so we combine them additively:
\begin{equation}
z_\theta(x)=z_0(x)+\Delta_\theta(x),
\label{eq:shared-representation}
\end{equation}
where $\Delta_\theta$ adds clock-aware information in the same representation space. Starting this update at zero preserves the initial content path; histories without sufficient support retain the base. TER trains $z_0$ and then $\Delta_\theta$ with the same evidence objective. At inference, the encoder outputs one $z_\theta$ shared by all tasks; neither evidence targets nor episodic regressors are required.

\subsection{Training and downstream use}

Training proceeds in three steps (Figure~\ref{fig:architecture}): local self-supervision learns the daily encoder; TER then learns multi-day context; finally, TER learns the clock-aware memory while keeping the preceding modules unchanged. For the final model, the daily and dynamics branches are adapted on a larger unlabeled pool and combined with the trained longitudinal modules, without joint retraining. Episodic Ridge regressors are discarded; downstream tasks fit separate linear classifiers (Appendix~\ref{sec:multistage-contract}).

\section{Experiments}
\label{sec:experiments}

Our comprehensive representation evaluation covers GlucoFM's complete public phenotype suite \citep{li2026glucofm}, private-cohort meal responses, and public-trial future outcomes. We then examine direct evidence, joint reconstruction, and simple long-history encoding.

\subsection{Datasets and evaluation}

Pretraining uses 23,336 distinct private and 2,283 public unlabeled 24-hour windows, excluding identified downstream participants in Shanghai and Stanford; CGMacros and Hall are absent. No phenotype labels enter representation learning. The public-only ablation retrains all stages on public windows. Appendix~\ref{sec:data-inventory} documents data, identity boundaries, and authorization. The supplement provides checkpoints and code for extraction, evaluation, and public-only training.

We include all four cohorts and 14 cohort--task pairs in GlucoFM's phenotype benchmark: CGMacros, ShanghaiT2DM, Stanford, and Hall \citep{das2025cgmacros,zhao2023chinese,metwally2025subphenotypes,hall2018glucotypes}. Its seven phenotype families are diabetes, insulin resistance, $\beta$-cell dysfunction, hypoglycemia, glucotype, obesity, and hyperlipidemia. They provide 855/609/994/250 windows from 45/58/37/56 participants, respectively (label definitions in Appendix~\ref{sec:task-evidence-map}).

TER uses up to seven days of causal history (four in Hall); baselines retain their native input interfaces. Table~\ref{tab:main} compares complete methods, and Section~\ref{sec:history-use} tests long-context use with same-history controls.

Our method and three locally evaluated CGM comparators share ten repetitions of five-fold cross-validation. Candidate representations are selected only within the training portion of each outer fold using inner validation, with one shared candidate for all 14 tasks; the selected candidate is then evaluated once on the held-out outer identities. Within every fit, StandardScaler and L2 logistic regression are estimated from the training rows and applied to the held-out rows. PR-AUC, ROC-AUC, and Macro-F1 are averaged equally over 14 tasks (Appendix~\ref{sec:model-selection}). All recordings of the same participant remain together, including Shanghai's 65 visits from 58 people.

The CGM baselines span masked glucose reconstruction (CGMformer), JEPA-style latent prediction (CGM-JEPA; X-CGM-JEPA also predicts glucodensity), and masked-context/state--event prediction (GlucoFM) \citep{lu2025cgmformer,muhammad2026cgmjepa,li2026glucofm}. Six generic time-series models add breadth. For GlucoFM, Table~\ref{tab:main} identifies the paper-reported private-data result separately because its pretraining data and trained checkpoint are not released. Same-fold comparisons use released CGM checkpoints and our GlucoFM reproductions.

\begin{table}[!b]
\centering
\caption{Main results on four CGM cohorts. PR/AUC/F1 denote PR-AUC, ROC-AUC, and Macro-F1 (\%); Overall averages 14 tasks equally. Bold highlights column-best results, emphasizing TER for ties at the displayed precision. Public-only rows are unbolded.}
\label{tab:main}
\fontsize{7.7}{9.0}\selectfont
\setlength{\tabcolsep}{1.4pt}
\renewcommand{\arraystretch}{1.16}
\begin{tabular*}{\textwidth}{@{\extracolsep{\fill}}l*{15}{r}@{}}
\toprule
\multirow{2}{*}{Method}
& \multicolumn{3}{c}{CGMacros (4 tasks)}
& \multicolumn{3}{c}{ShanghaiT2DM (3)}
& \multicolumn{3}{c}{Stanford (3)}
& \multicolumn{3}{c}{Hall (4)}
& \multicolumn{3}{c}{Overall (14 tasks)} \\
\cmidrule(lr){2-4}\cmidrule(lr){5-7}\cmidrule(lr){8-10}\cmidrule(lr){11-13}\cmidrule(lr){14-16}
& PR & AUC & F1 & PR & AUC & F1 & PR & AUC & F1 & PR & AUC & F1 & PR & AUC & F1 \\
\midrule
\multicolumn{16}{l}{\emph{Strong generic time-series foundation models: reported results$^\dagger$}} \\
Chronos-2 (small)$^\dagger$ & 55.2 & 60.3 & 53.5 & 38.6 & 51.6 & 51.5 & 63.4 & 62.3 & 58.6 & 43.8 & 61.6 & 56.8 & 50.1 & 59.2 & 55.1 \\
Chronos-2 (120M)$^\dagger$ & 57.6 & 63.0 & 55.4 & 37.8 & 50.7 & 50.7 & 62.1 & 61.2 & 58.0 & 43.7 & 59.1 & 55.9 & 50.4 & 58.9 & 55.1 \\
MOMENT (small)$^\dagger$ & 57.3 & 62.5 & 55.0 & 37.6 & 52.6 & 51.2 & 63.5 & 62.7 & 58.8 & 46.8 & 60.6 & 58.3 & 51.4 & 59.9 & 55.9 \\
MOMENT (large)$^\dagger$ & 57.9 & 63.6 & 55.7 & 38.8 & 52.7 & \textbf{51.8} & 63.9 & 62.5 & 58.5 & 42.8 & 59.1 & 54.8 & 50.8 & 59.7 & 55.2 \\
Mantis$^\dagger$ & 61.6 & 66.6 & 58.7 & 37.6 & 50.9 & 49.9 & 68.6 & 67.0 & 62.2 & 53.1 & 68.5 & 61.6 & 55.5 & 63.9 & 58.4 \\
MantisV2$^\dagger$ & 62.0 & 66.6 & 57.6 & 39.8 & 53.6 & 50.9 & 68.5 & 67.0 & 61.6 & 55.9 & 69.4 & 62.8 & 56.9 & 64.7 & 58.5 \\
\midrule
\multicolumn{16}{l}{\emph{CGM-specific baselines: released checkpoints and a reported private-data result$^\ddagger$}} \\
CGMformer & 61.9 & 69.2 & 57.7 & 37.4 & 52.4 & 48.2 & 64.9 & 64.5 & 56.2 & 50.7 & 62.9 & 59.4 & 54.13 & 62.80 & 55.82 \\
CGM-JEPA & 62.1 & 65.4 & 55.5 & 40.4 & 54.3 & 47.3 & 65.4 & 64.7 & 55.7 & 55.8 & 66.4 & 62.7 & 56.35 & 63.15 & 55.84 \\
X-CGM-JEPA & 62.4 & 66.0 & 55.8 & 39.6 & 53.0 & 47.2 & 65.8 & 65.5 & 56.4 & 56.2 & 66.3 & 63.1 & 56.47 & 63.20 & 56.18 \\
GlucoFM (reported)$^\ddagger$ & 64.7 & 69.3 & 59.4 & 40.5 & 55.8 & 51.7 & 71.3 & 70.2 & \textbf{64.5} & 57.3 & 69.7 & 63.0 & 58.8 & 66.7 & 59.9 \\
\midrule
\multicolumn{16}{l}{\emph{Private+public unlabeled pretraining; common downstream folds}} \\
GlucoFM (repro., private+public) & 62.2 & 67.1 & 56.5 & 38.3 & 51.7 & 46.0 & 69.6 & 71.5 & 59.6 & 46.2 & 58.6 & 55.1 & 54.10 & 62.34 & 54.50 \\
\textbf{Ours (TER)} & \textbf{70.9} & \textbf{75.1} & \textbf{64.1} & \textbf{45.0} & \textbf{56.4} & 46.9 & \textbf{71.6} & \textbf{73.5} & 61.6 & \textbf{64.7} & \textbf{76.4} & \textbf{66.4} & \textbf{63.75} & \textbf{71.13} & \textbf{60.56} \\
\midrule
\multicolumn{16}{l}{\emph{Matched public-only pretraining: identical corpus and downstream folds}} \\
GlucoFM (repro.) & 61.2 & 66.8 & 56.9 & 36.4 & 48.1 & 45.6 & 69.9 & 72.7 & 60.8 & 42.4 & 56.6 & 52.1 & 52.39 & 61.11 & 53.96 \\
Ours (public-only TER) & 68.8 & 74.6 & 61.6 & 38.2 & 50.8 & 42.8 & 69.8 & 71.4 & 60.5 & 56.3 & 70.5 & 61.3 & 58.88 & 67.63 & 57.25 \\
\bottomrule
\end{tabular*}
\par\smallskip
\begin{minipage}{\linewidth}\footnotesize
$^\dagger$Values for the six generic foundation models are transcribed from the GlucoFM study and retain its 10$\times$5 grouped-fold linear-probe protocol \citep{li2026glucofm}. $^\ddagger$The reported GlucoFM model was pretrained on non-public Wear-CGM data and is cited from the paper because its checkpoint is unavailable. The other three CGM rows evaluate released checkpoints on our folds and probe. The public-only TER--GlucoFM pair shares both pretraining data (2,283 windows) and downstream evaluation. Differences quoted in text use unrounded scores.
\end{minipage}
\end{table}

\subsection{Main results on 14 phenotype tasks}
\label{sec:results}

Table~\ref{tab:main} compares TER with ten leading CGM and time-series baselines on 14 phenotype tasks across four cohorts. TER leads in \textbf{\finalcells{}/42 metric cells across \finaltasks{}/14 tasks}. A task is counted when TER achieves at least one new best metric. Overall PR-AUC, ROC-AUC, and Macro-F1 are \finalpr\%, \finalauc\%, and \finalf\%. A separate eight-chain training study yields sample standard deviations of 0.303/0.326/0.239 percentage points (Appendix~\ref{sec:multistage-contract}).

\paragraph{Magnitude and breadth of the gains.}
TER exceeds GlucoFM, the strongest prior model on the three Overall metrics, by \textbf{4.95/4.43/0.66 points}. Its PR-AUC and ROC-AUC gains are 2.6 and 2.2 times the respective gaps between GlucoFM and MantisV2, the next strongest baseline. TER leads 5 of seven phenotype families in PR-AUC (Appendix Figure~\ref{fig:phenotype-family-breadth}). Appendix~\ref{sec:benchmark-details} reports individual tasks.

\paragraph{Data-matched public-only comparison.}
To test whether the advantage depends on private pretraining data, we train both methods on the same 2,283 public windows and use identical downstream folds and classifiers. TER improves our GlucoFM reproduction by \textbf{6.50/6.52/3.30 points} and wins 30/42 paired task--metric cells. The improvement thus persists when both methods learn from the same available corpus.

\paragraph{Publicly available CGM checkpoints on common folds.}
Released CGMformer, CGM-JEPA, and X-CGM-JEPA checkpoints share our rows, folds, and linear probe. TER exceeds the strongest on each Overall metric by \textbf{7.29/7.93/4.38 points} (Appendix Figure~\ref{fig:checkpoint-gap}). This controls downstream evaluation; the public-only comparison additionally controls pretraining data.

\subsection{Transfer to meal responses and future glucose outcomes}
\label{sec:ppgr-main}
\begin{figure}[H]\centering
\includegraphics[width=.94\linewidth]{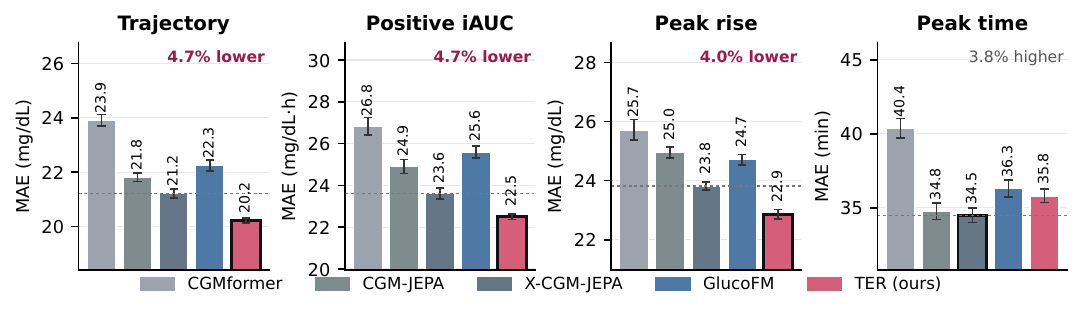}
\caption{\textbf{Daily representation-only transfer to postprandial prediction.} Subject-macro MAE (mean $\pm$ SD across ten common-head initializations); lower is better. All models use only preceding 24-hour CGM; GlucoFM is our reproduction.}
\label{fig:ppgr-transfer}
\end{figure}

Two additional datasets test subsequent responses and future outcomes using daily and longitudinal representations, respectively.
\paragraph{Daily representations for postprandial prediction.}
We test the daily encoder on a private meal-response dataset containing 1,705 meals from 409 people. Each model supplies a representation of the preceding 24-hour CGM to common prediction heads, with no meal or outcome labels used in pretraining. TER has the lowest trajectory, positive-iAUC, and peak-rise errors among five CGM models (Figure~\ref{fig:ppgr-transfer}), improving the strongest competing means by \textbf{4.70\%, 4.71\%, and 4.02\%}. Its peak-time error is 3.81\% higher than X-CGM-JEPA's. These results show that the daily representation retains information about the magnitude and course of a subsequent response. Appendix~\ref{sec:ppgr-transfer} evaluates added meal context and identity-disjoint sensitivity.

\paragraph{Longitudinal representations for future CGM outcomes.}
\label{sec:aide-main}

The public AIDE T1D trial \citep{kudva2025aide} tests longitudinal use in new participants. Four to seven observed days from 81 participants (236 treatment periods) predict seven CGM outcomes in a later, non-overlapping period. TER, a label-free BiLSTM, and mean--max pooling receive the \emph{same daily representations, histories, and treatment covariate} under participant-grouped folds. TER has lower error on \textbf{all seven} outcomes; for time below 70 mg/dL, MAE falls by \textbf{39.1\%} versus BiLSTM and \textbf{47.4\%} versus mean--max. This extends the evidence for learned longitudinal aggregation to a new cohort and a forward-looking task. Appendix~\ref{sec:aide-external} gives the complete protocol, outcomes, and separate held-out evaluation.

\subsection{Why learn from evidence rather than export it directly?}
\label{sec:learning-audit}

\emph{Direct summaries expose sample-specific associations alongside useful signal.} We test whether their representation contributes to the deficit without adding information. An affine map learned from 7,341 unlabeled windows maps daily-plus-evidence inputs to TER coordinates. The same downstream classifier gains 4.87/4.97/1.38 PR-AUC/ROC-AUC/Macro-F1 points, with no additional observations or nonlinear decision boundary. Part of the deficit therefore lies in how existing information is weighted by a finite-sample predictor (Appendix~\ref{sec:summary-projection}).

Figure~\ref{fig:feature-transfer} locates the problem: some class differences in the fitting group fail to recur in held-out groups. Evidence can supervise a learned representation instead of being exported directly; the next question is how to train its recovery.

\subsection{Why not jointly learn an evidence decoder?}
\label{sec:why}

\emph{Accurate recovery and a stable decoding rule are different requirements.} Joint training optimizes one encoder--decoder pair; TER fits the rule on one group and scores another. We separate target choice from this training procedure, then isolate cross-group scoring.

We cross structured evidence and raw-hourly targets with joint linear decoding and episodic fit--transfer. All four share the final daily input, 8,421-window unlabeled corpus, budgets, assembly, and evaluation. Each arm repeats training eight times at the same fixed hyperparameters and seed; training-internal CV selects one shared representation per outer fold; Appendix~\ref{sec:objective-isolation} gives the training and selection protocol.

\begin{figure}[H]\centering
\includegraphics[width=.97\linewidth]{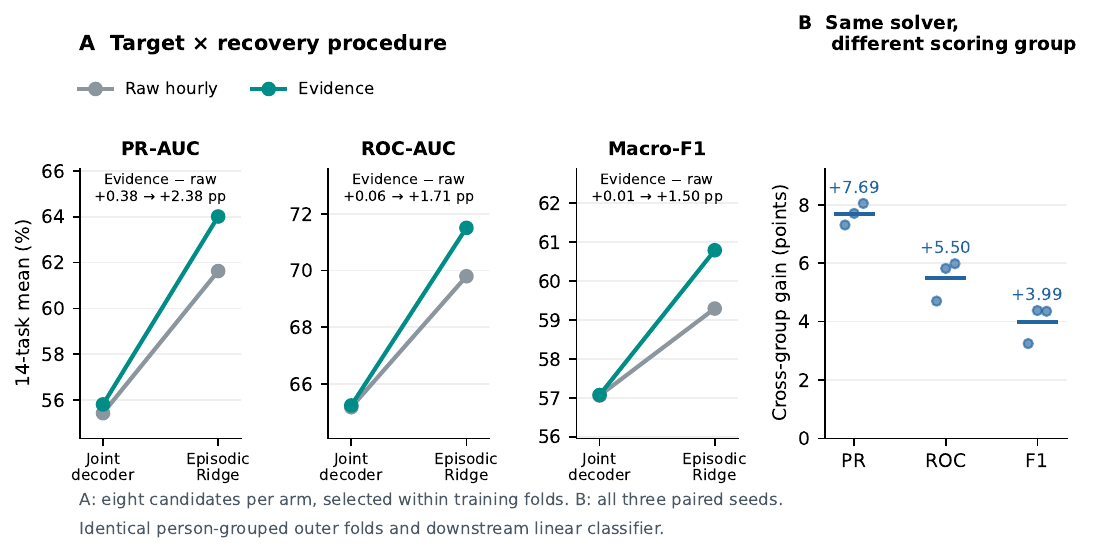}
\caption{\textbf{Target choice and cross-group recovery.} \textbf{A:} matched target $\times$ recovery ablation with eight candidates per arm. \textbf{B:} changing only the scoring group while keeping the episodic Ridge solver fixed. Points show all three paired seeds; bars and annotations show their mean gains.}
\label{fig:canonical-main-why}
\end{figure}

Averaged over both targets, fit--transfer adds \textbf{7.20/5.45/2.97 points} in PR-AUC/ROC-AUC/Macro-F1 (Figure~\ref{fig:canonical-main-why}A). Structured evidence further improves fit--transfer over raw recovery. The target helps, but changing the recovery procedure drives the larger gain.

Three paired-seed interventions isolate the cross-group requirement: both arms solve the same two Ridge regressors per update, changing only scoring on fitting people versus the opposite group. Cross-group scoring adds 7.69/5.50/3.99 PR-AUC/ROC-AUC/Macro-F1 points and reduces fit-group prediction variance by 62.6--67.1\%, while average held-out evidence error changes little. The gain accompanies reduced dependence on who fitted the rule, rather than closer average evidence recovery (Appendix~\ref{sec:fit-transfer-intervention}).

\paragraph{Cross-person consistency of class differences.}
To locate sample dependence in the features, we keep each encoder fixed and use the same outer folds and inner-selected encoders as the primary comparison. On standardized training features only, PCA identifies the leading combinations explaining at least 90\% of feature variance; we examine the remaining, smaller variations. For each class, we subtract the mean feature vector of the other classes, separately for fitting and held-out groups. Panel~A measures whether these class-difference vectors point in the same direction: cosine near zero means little agreement. Panel~B removes only these directions' contribution to the fitted classifier's prediction, without retraining it. Higher PR-AUC means that contribution was harmful on balance; lower PR-AUC means it carried useful signal. No held-out data select the directions (Appendix~\ref{sec:feature-transfer}).

\begin{figure}[H]\centering
\includegraphics[width=\linewidth]{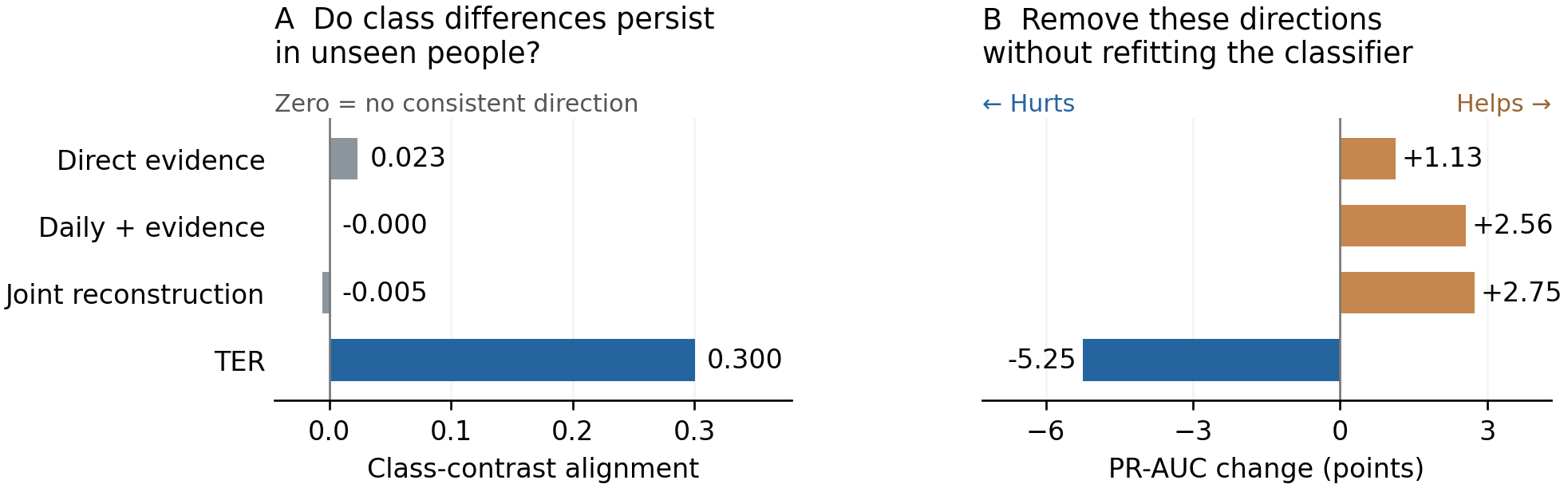}
\caption{\textbf{Which feature differences survive a change of people?} \textbf{A:} cross-group agreement of class differences in the smaller-variance directions. \textbf{B:} PR-AUC change after removing their contribution from the same classifier. Results average all 14 tasks. The shared variance rule selects different ranks; all seven arms and details are in Appendix~\ref{sec:feature-transfer}.}
\label{fig:feature-transfer}
\end{figure}

For direct evidence, concatenation, and joint reconstruction, these class differences show near-zero cross-group agreement; removing them improves PR-AUC by 1.13/2.56/2.75 points (Figure~\ref{fig:feature-transfer}). TER retains positive agreement; removal loses 5.25 points and hurts 13/14 tasks. Small variation is therefore not inherently noise: the problem is using differences that separate fitting samples but do not recur in new groups. Repeated evidence-readout fits also show reduced fitting-group sensitivity after output-scale normalization (Appendices~\ref{sec:readout-mechanism} and~\ref{sec:contrast-retention}).

\subsection{Why not simply pool or encode a longer history?}
\label{sec:history-use}

The transfer objective needs the relevant history to survive encoding. Mean--max removes day order; BiLSTM can retain it, but daily reconstruction rewards input fidelity rather than explicitly targeting recurrence or across-day change. TER targets these properties, while clock-aware memory retains value--time relations.

With identical daily inputs and histories, the fixed-configuration TER component study exceeds mean--max and BiLSTM by 8.67 and 9.24 PR-AUC points. Seven-day raw encoding with 38,527 updates reaches 57.46\% versus TER\textquotesingle s 63.75\%. Reassigning value--time rows costs 1.64 PR-AUC points in the matched component study: organization matters beyond availability (Appendices~\ref{sec:components} and~\ref{sec:label-history-transfer}).

With BiLSTM, inputs, initialization, and budget fixed, changing daily reconstruction to evidence fit--transfer reduces across-day mean/change/event readout errors by 24.1/17.1/22.8\% and adds 0.42/0.24/0.39 phenotype metric points. This partial repair demonstrates the role of the learning objective, rather than recurrent incapacity: longitudinal properties receive explicit supervision (Appendix~\ref{sec:bilstm-objective-repair}).

\section{Conclusion}

TER turns longer histories into transferable evidence through clock-aware encoding and cross-group recovery. Matched interventions connect its gains to reduced fitting-group sensitivity; phenotype and future-outcome evaluations establish predictive utility.
\label{main:last}

\clearpage
\section*{AI use statement}
Generative AI assisted conceptual formulation, literature inspection, mathematical exposition, figure and manuscript drafting, artifact checks, and orchestration of the prespecified objective controls. Numerical claims were checked against existing experiment reports. AI assisted implementation of the specified synthetic timing experiment; its signals follow a deterministic generator, not model-generated examples. No phenotype labels entered representation training. The authors remain responsible for the research, citations, and final submission.

\section*{Reproducibility statement}
Sections~\ref{sec:formulation}--\ref{sec:experiments} specify the learning objective, encoder, and evaluation. The appendices define the evidence, normalization, training stages, and experiments. The code supplement provides a deployment checkpoint, all 24 main-table candidates with shared daily components, nested-evaluation code, fold and selection records, and public-only training scripts. Its README distinguishes checkpoint inference from main-table reproduction.

\section*{Ethics statement}
This retrospective representation study does not validate clinical diagnosis or treatment. Cohort size, heterogeneous acquisition, and nonuniform insulin-related performance limit medical interpretation. Identifiers are used for grouping rather than as model features; grouping alone does not establish privacy. Private CGM records were de-identified and used with user authorization for research; raw or person-level private data are not released. No new recruitment, intervention, or collection occurred, and original access and redistribution conditions apply.

\clearpage
\label{references:first}
\bibliography{references_probe_transfer}
\bibliographystyle{plainnat}
\clearpage
\appendix
\section{Transferable evidence reconstruction: targets and episodic Ridge regression}
\label{sec:target-details}
\begin{table}[H]\centering
\caption{\textbf{Complete inventory of the evidence $e=H(x)$.} The ten rows expand the three evidence families summarized in the main text; their counts sum to 336 scalar evidence values. All glucose thresholds are in mg/dL. These are signal summaries, not phenotype labels or independent physiological measurements.}
\label{tab:targets}
\small\setlength{\tabcolsep}{4pt}\renewcommand{\arraystretch}{1.12}
\begin{tabular}{@{}>{\raggedright\arraybackslash}p{.18\linewidth}r>{\raggedright\arraybackslash}p{.70\linewidth}@{}}
\toprule
Component & Count & Exact contents / construction \\\midrule
\multicolumn{3}{@{}l}{\textit{Current-day evidence: four groups of sixteen}}\\
State & 16 & Mean and standard deviation (2); quantiles .05/.10/.25/.50/.75/.90/.95 (7); fractions $<70$, $70$--$180$ inclusive, $>180$ (3); observed means in recorded 00--06, 06--10, 10--18, 18--24 hour intervals (4).\\
Rise/recovery excursions & 16 & Proposal count, capped at 6 and divided by 6 (1); the top three proposals each contribute 60-minute rise, peak amplitude, time to peak in hours, and glucose drop 60/120 minutes after the peak ($3\times5$).\\
Extremes & 16 & Fractions $<70$, $<54$, $>180$, $>250$ (4); minimum, maximum, q05, q95 (4); low/high run counts and longest durations in minutes (4); maximum 30-minute decline/rise (2); largest 120-minute endpoint rebound/recovery after low/high observations (2).\\
Rhythm / observation & 16 & Six consecutive four-hour means from the window start (6); observed-weighted sine/cosine coefficients at daily harmonics 1 and 2 (4); standard deviation, last--first change, and mean absolute adjacent change of the six means (3); 60-minute lag correlation (1); cadence-adjusted coverage and cadence/15 minutes (2).\\\midrule
\multicolumn{3}{@{}l}{\textit{Longitudinal evidence: four summaries of all 64 current-day targets}}\\
Mean & 64 & For each coordinate, $\mu=\sum_d p_d x_d$, using positive family-support weights normalized over observed history.\\
Spread & 64 & For each coordinate, $\sigma=\sqrt{\sum_d p_d(x_d-\mu)^2}$.\\
Directed extreme & 64 & Maximum over supported days; minimum instead for the current extreme group's minimum and q05 coordinates.\\
Ordered change & 64 & Most recent minus oldest supported value, $x_{\mathrm{last}}-x_{\mathrm{first}}$.\\\midrule
\multicolumn{3}{@{}l}{\textit{Low/high event evidence: eight descriptors per event type}}\\
Low glucose & 8 & From observed daily fractions $b_d$ below 70: weighted occurrence prevalence; occurrence-day count/7; adjacent-day persistence, onset and offset; current occurrence; weighted mean burden; weighted burden spread. Occurrence means $b_d>0$.\\
High glucose & 8 & The same eight descriptors, using observed daily fractions above 180. Adjacency requires exactly 24 hours; onset/offset are oriented from past to present.\\
\bottomrule\end{tabular}
\end{table}

\subsection{Current evidence: four families of sixteen}
The current bank contains four evidence families, each with sixteen coordinates. They describe the observed signal and its acquisition; they are not independent diagnoses or direct measurements of latent physiology.

\paragraph{State.}
Table~\ref{tab:targets} lists the sixteen coordinates in their exported order. They use observed values only; a clock interval with no observations has mean zero. Clock intervals use the recorded timestamp, not an inferred sleep or fasting annotation.

\paragraph{Challenge.}
A proposal requires an observed start and 60-minute endpoint, a rise of at least 30 mg/dL, and at least half the cadence-expected coverage in between. Its peak is the largest observed value in the following up-to-120-minute interval. Proposals are ranked by peak amplitude plus the positive part of 120-minute recovery, and the top three are kept without an additional deduplication step. Missing proposals are zero-padded; recovery endpoints are clipped to the window boundary. These are candidate signal excursions, not identified meals; no meal annotations are used.

\paragraph{Extreme.}
Low/high runs use thresholds 70/180 and join successive observed readings only when their separation is at most $\max(5,\text{cadence})$ minutes. The longest run duration is its observed count times this interval. Decline/rise uses 30-minute differences of the filled profile. Rebound/recovery compares each qualifying observation with the value 120 minutes later, clipped to the window end, then takes the maximum; it is not the maximum over all intervening times.

\paragraph{Rhythm.}
The six block means follow the window start, whereas the harmonic phases use recorded clock time. Harmonics use the filled signal centered by its observed mean, divided by $\max(\text{observed standard deviation},1)$, and weighted at observed positions. The lag correlation uses the filled profile. Filling is linear interpolation with endpoint extension; nonfinite descriptors become zero. This target preprocessing differs from the gap-safe raw pairs used by clock-aware memory.

Support for state, extreme, and rhythm reflects coverage corrected for expected cadence. Challenge support combines coverage with the availability of up to three proposals. These fixed domain rules are part of the inductive bias; absence of phenotype labels does not make the bank knowledge-free.

\subsection{Longitudinal and event evidence}
For each evidence family, retain current/history days having positive family support. With normalized support weights $p_d$ and chronological order oldest to newest, each coordinate contributes
\[
\mu=\sum_d p_dx_d,\qquad \sigma=\sqrt{\sum_d p_d(x_d-\mu)^2},
\qquad T=\max_d x_d,\qquad \Delta=x_{\mathrm{last}}-x_{\mathrm{first}}.
\]
Only the extreme family's minimum and q05 coordinates use a minimum for $T$. Thus the directed extreme is not a common quantile operator across all coordinates. Flattening family, functional, and coordinate produces $4\times4\times16=256$ values after the current bank. Missing family support yields zeros; a single supported day yields zero spread and change.

Events use observed low ($<70$ mg/dL) and high ($>180$ mg/dL) daily burdens. Each contributes eight descriptors: coverage-weighted prevalence, number of occurrence days divided by seven, adjacent-day persistence, past-to-present onset, past-to-present offset, current occurrence, mean burden, and burden variability. Adjacent days must be exactly 1,440 minutes apart; missing days are not made adjacent by interpolation. Pair weight is the smaller coverage. Variability uses $\sqrt{\max(v,0)+10^{-8}}-10^{-4}$ so a constant sequence gives exactly zero. The order is low-eight then high-eight.

The first 320 evidence coordinates use center and scale estimated once from eligible pretraining fitting rows, with standard-deviation floor $10^{-5}$; the 16 event coordinates use the context encoder's fitting-split event transform. These fixed transforms use no audit rows or downstream phenotype annotations. They may include rows later sampled into either episode role. Separately, each episode estimates its feature scaler and regression intercept only from the fit group.

\paragraph{Main targets versus auxiliary targets.}
The 336 evidence values summarized in the main text and Table~\ref{tab:targets} form the main objective in both multi-day training stages. Context training additionally uses a separate episode to recover the sixteen state coordinates of the already observed final day, not sixteen new entries appended to $H$. The final memory stage omits that auxiliary episode (Appendix~\ref{sec:multistage-contract}). The spectral recoverability diagnostic below is also separate and never contributes optimizer gradients.

\subsection{Sampling and numerics}
\label{sec:ridge}
\paragraph{Differentiable Ridge regression.}
Let $Z_F,Z_T$ denote representations and $E_F,E_T$ their fixed evidence. Estimate representation mean and scale on $F$ alone to obtain $P_F,P_T$. With evidence mean $\bar e_F$, the fitted regressor is
\begin{align}
B_F&=\arg\min_B\|P_FB-(E_F-\mathbf1\bar e_F)\|_F^2+\|B\|_F^2,\nonumber\\
\widehat E_T&=P_TP_F^\top\operatorname{solve}
(P_FP_F^\top+I,\;E_F-\mathbf1\bar e_F)+\mathbf1\bar e_F .
\label{eq:ridge}
\end{align}
The reconstruction loss differentiates through both the fitting and application paths:
\begin{equation}
\nabla_\psi\mathcal L
=J_\psi z_\psi(F)^\top\frac{\partial\mathcal L}{\partial Z_F}
+J_\psi z_\psi(T)^\top\frac{\partial\mathcal L}{\partial Z_T}.
\label{eq:gradient}
\end{equation}
Here $\psi$ denotes the active encoder parameters; it corresponds to $\theta$ in the main objective. The two roles use the same encoder.

\paragraph{Linear inner--outer intuition.}
Suppose a centered signal has covariance $\Sigma$, a linearized representation is $z=\Lambda^\top x$, and linear evidence is $e=U^\top x$. The population Ridge regressor for a fixed representation is
\begin{equation}
W^\star(\Lambda)=
(\Lambda^\top\Sigma\Lambda+\lambda I)^{-1}\Lambda^\top\Sigma U.
\label{eq:linear-inner-reader}
\end{equation}
The inner fit therefore exposes the encoder only through directions supported jointly by signal covariance and the evidence mapping $\Sigma U$. TER repeatedly re-estimates the finite-sample analogue of Equation~\ref{eq:linear-inner-reader} and scores it on disjoint identities; a joint decoder instead carries its own decoding state across updates. This calculation motivates the fit--transfer bias but is not a theorem that episodic fitting dominates joint decoding for arbitrary nonlinear encoders. The matched ablation in Appendix~\ref{sec:objective-isolation} measures the effect in the final training pipeline.

Endpoints need at least one prior window for fitting/audit preparation. Fixed pretraining fitting and audit partitions use recorded identities. Within the fitting partition, an episode shuffles identities into disjoint fit and transfer groups, then samples 64 rows per side with replacement by choosing an identity uniformly and one of its available endpoints uniformly. At least four fit-group identities are required. Rows with fewer than three supported days may enter an episode with their clock-aware memory gate closed. Identities can exchange roles in later episodes.

The Ridge solve uses float64. For fit-group population variance $v_r$, the feature scale is one when $v_r=0$ and $\sqrt{\max(v_r,10^{-10})}$ otherwise. The fit-group evidence mean supplies the unpenalized intercept. The sum-squared-error regularizer in Equation~\ref{eq:ridge} is one; it would be $1/64$ if the fit loss were divided by fit-group size. The solve result is returned to representation dtype for the outer loss. Neither the fit-group scaler nor the fitted regressor is detached.

The outer loss gives equal weight to current-day evidence and the remaining coordinates:
\begin{equation}
\mathcal L_{F,T}=
\frac{\|\widehat E_T^{1:64}-E_T^{1:64}\|_F^2}{64|T|}
+\frac{\|\widehat E_T^{65:336}-E_T^{65:336}\|_F^2}{272|T|}.
\label{eq:outer}
\end{equation}

Each episode samples disjoint recorded-identity groups and rows, encodes both sides with shared parameters, derives fixed evidence, estimates fit-group-only feature scaling and Ridge regression, and computes the two-block transfer loss without refitting. Backpropagation passes through both representations and the episodic Ridge fit before clipping and AdamW. The fitted regressor is discarded after the update. Each clock-aware memory run exports its final-step checkpoint; selection across runs follows Appendix~\ref{sec:model-selection}.

\subsection{Evidence-target and recovery controls}
\label{sec:objective-isolation}
\label{sec:final-assembly-bridge}
The four arms cross structured versus raw-hourly targets with joint linear decoding versus episodic Ridge recovery. They use the same daily input, 8,421-window corpus, episode sequence, seed 43, learning rate $4\times10^{-4}$, auxiliary weight 1.75, 6,960 context updates, 10,440 memory updates, and final assembly. Each arm repeats training eight times to capture accelerator numerical variation. We select among its eight runs by the inner-CV rule in Section~\ref{sec:model-selection}, then score the common outer folds. The matched evidence arm restricts TER to this same configuration; the main table uses the full 24-candidate pool.

Joint-decoder controls initialize affine evidence heads by Ridge, then optimize the heads jointly with the encoder. The context stage has main and observed-last-day auxiliary heads; the memory stage has only the main head. TER instead solves its regressor in each episode and differentiates the other-group loss through the solve. Both branches retain the same observation-supported inputs and training-only evidence.

Raw-hourly targets right-align seven days into 168 hourly bins, apply $\tanh((g-100)/40)$ to observed hourly means, and append 168 observation indicators. Empty hours have zero value and mask. Structured targets use the 336 properties defined above. Target families retain equal aggregate weighting; phenotype labels never enter these objectives.

\begin{table}[H]\centering\small
\caption{Target $\times$ recovery ablation under equal candidate budgets. Each arm uses eight repeated runs at the same configuration, selected by inner CV; the final row also gives the main-table 24-candidate result. Scores are percentages.}
\label{tab:canonical-factorial-absolute}
\setlength{\tabcolsep}{4pt}
\begin{tabular*}{\linewidth}{@{\extracolsep{\fill}}lrrr@{}}
\toprule
Target and recovery & PR-AUC & ROC-AUC & Macro-F1\\\midrule
Raw hourly, joint decoder & 55.43 & 65.17 & 57.06\\
Evidence, joint decoder & 55.81 & 65.24 & 57.07\\
Raw hourly, episodic Ridge & 61.63 & 69.80 & 59.29\\
Evidence, episodic Ridge (matched eight) & 64.01 & 71.51 & 60.79\\
TER (main-table selection) & 63.75 & 71.13 & 60.56\\
\bottomrule\end{tabular*}
\end{table}

The target main effect is 1.38/0.89/0.75 percentage points; the recovery-procedure main effect is 7.20/5.45/2.97; their difference-in-differences is 2.00/1.65/1.48. These are the paired outer-test effects of the equal-budget four-arm comparison.

\subsection{Recorded identity and source boundaries}
\label{sec:identity-scope}
The 8,421-window corpus used by the matched TER and full-pipeline joint-decoder variants includes 6,138 unlabeled windows from a private CGM source and 2,283 public windows (Appendix~\ref{sec:data-inventory}). It contains 7,341 fitting and 1,080 audit windows grouped by stored identities; another 649 held-out private-source windows are excluded. Both variants use the same scaled Stage-A checkpoints documented in Appendix~\ref{sec:data-inventory}, with 6,960 context-stage and 10,440 memory-stage updates. Public sources are BIG IDEAs, Col\'as, ShanghaiT2DM, and Stanford \citep{cho2023bigideas,colas2019dfa,zhao2023chinese,metwally2025subphenotypes}. Both TER episode roles draw only from eligible fitting rows, independently of downstream folds. The fixed audit partition is outside optimizer updates but repeatedly used for diagnostics. Global evidence transforms may see records later drawn into either role; only the episode's feature scaler, intercept, and Ridge fit use the fit group.

Shanghai pretraining keys identify recording visits. The saved public audit retains 43 visits from 42 biological participants, including two visits of one participant; both enter the fitting partition under the saved construction rule. They can occupy opposite episode roles under the implemented recorded-identity sampler. Actual historical episode overlap has not been replayed. This differs from downstream exclusion: the source audit excludes downstream Shanghai biological participants, including alternate visits, and downstream Stanford participants. Corpus construction preserves the audited public rows when removing the private source. Saved audits and source hashes do not establish global biological-person linkage across independent sources.

The downstream evaluation groups Shanghai's 65 visits by their 58 biological participants. All visits of one person remain in the same outer or inner fold. The locally evaluated models use this common person-grouped plan; GlucoFM's reported row retains the source paper's results.

\subsection{Training, model selection, and evaluation}
\label{sec:model-selection}
Representation learning uses unlabeled CGM and signal-derived evidence. Identified downstream participants, including alternate Shanghai visits, are excluded from the pretraining corpus. The phenotype evaluation uses ten repetitions of five biological-person-grouped outer folds, shared across tasks within each cohort. StandardScaler and L2 logistic regression ($C=1$, LBFGS, at most 1,000 iterations) are fitted only on the relevant training rows.

TER has 24 candidates: eight repeated runs at each of three context-stage learning-rate/auxiliary-weight pairs, $(3.5\times10^{-4},1.5)$, $(4\times10^{-4},1.75)$, and $(5\times10^{-4},1.0)$. The memory-stage setting is shared. In each outer training population, three inner folds evaluate every candidate. The equal-task mean of PR-AUC, ROC-AUC and Macro-F1 selects one candidate shared by all 14 tasks. Its classifiers are refitted on the outer training data and scored on the held-out outer identities. Exact ties use candidate order. The main table averages these outer-test results, not an outer-score-selected checkpoint.

Partitions balance person counts and require class coverage for each task. Inner partitions contain only outer-training participants. Features and model scores do not determine the partitions. Local checkpoint baselines and fixed ablations use these same outer folds without an encoder-selection step. The target--recovery factorial uses eight candidates per arm at the matched configuration. All metrics, transformations and selected candidate identities are archived per fold. AIDE and meal-response experiments retain their separately specified prediction protocols.

\paragraph{Reproduction entry points.}
The code supplement's \texttt{evaluate\_nested.py} extracts all 24 supplied candidates, performs the training-internal selection, and reports the main-table outer scores. It includes candidate checksums, common row partitions, input fingerprints, and the 50 fold selections with aggregate metrics. \texttt{extract.py} uses the deployment checkpoint for new recordings; \texttt{evaluate\_14task.py} evaluates a fixed checkpoint on the same person-grouped outer folds. A fixed-checkpoint score and the selected-procedure score are distinct outputs of these documented entry points.

\section{Observation-aware and clock-aware CGM representation: implementation}
\label{sec:daily-details}
\subsection{Observation-aware daily encoding: implementation}
\label{sec:daily-implementation}
Observation-aware daily encoding uses a mask-conditioned lift with constant edge extension, not an orthogonal projector. Common and native views share common-visible normalization. Its 96 common channels and 32 residual channels use residual dilations 1, 2, 4, and 8. The signed residual enters once; upper/lower groups of 16 channels use raw-glucose tail weights, falling back to observed-point pooling when the corresponding tail is absent. Training masks 25\% of observations and uses
\begin{equation}
\mathcal L_{\mathrm{daily}}=\mathcal L_{\mathrm{obs}}+\mathcal L_{\mathrm{shape}}+0.2\mathcal L_{\mathrm{sev}}+0.2\mathcal L_{\mathrm{nest}}+0.2\mathcal L_{\mathrm{cad}}.
\end{equation}
Here every reconstruction term is a mean squared error over its stated support. $\mathcal L_{\mathrm{obs}}$ averages native- and deterministic-15-minute-view reconstruction of held-out values that are observable in the common view. $\mathcal L_{\mathrm{shape}}$ reconstructs the off-center signed excursion profiles at the .90/.95/.975 training quantiles, and $\mathcal L_{\mathrm{sev}}$ reconstructs the corresponding peak excess. $\mathcal L_{\mathrm{nest}}$ matches the predicted profile difference between adjacent quantile levels to the fixed target difference. $\mathcal L_{\mathrm{cad}}$ is the mean squared distance between $\ell_2$-normalized 96-dimensional common-branch codes from the native and deterministic-15-minute views. Unsupported profile/contrast entries do not enter their reductions.

The auxiliary dynamics encoder uses two identical four-block dilated convolutional branches (dilations 1/2/4/8, width 32). Each pools a masked token mean and signed first temporal moment to 16 coordinates. Averaging forward and time-reversed outputs makes the state branch even; taking half their difference makes the dynamics branch odd. Its loss is physical-trace denoising $+0.2$ times MSE on 12 state functions $+$ MSE on ten signed-increment functions $+$ unit variance penalties $+0.04$ off-diagonal covariance penalties. The state functions are level, spread, quantiles, tail fractions, autocorrelation, and difference energy; dynamics functions are signed increments and signed energies at lags 1/2/4/8/16. Only the 16-dimensional odd branch is retained in the final daily representation.
Tail targets use fit-split empirical levels .90/.95/.975; cosine cadence consistency acts only on common channels. Affine coordinates restore location $(\mu_x-c_{\mathrm{train}})/s_{\mathrm{train}}$ and scale $\log\max(\sigma_x/s_{\mathrm{train}},10^{-6})$, with training glucose median $c_{\mathrm{train}}$ and standard deviation $s_{\mathrm{train}}$. The daily representation is ordered as
\begin{equation}
z_d=[z_o^{96},g_{\mathrm{affine}}^2,z_+^{16},z_-^{16},z_{\mathrm{odd}}^{16}]\in\mathbb R^{146},
\qquad z_{\mathrm{odd}}(x)=[f(x)-f(\operatorname{rev}x)]/2.
\label{eq:daily}
\end{equation}
The context encoder's output map $R:\mathbb R^{36}\!\to\!\mathbb R^{146}$ maps state corrections into these features. The clock-aware memory's zero-initialized $W:\mathbb R^{16}\!\to\!\mathbb R^{36}$ writes through $R$. Export reverses the fixed daily standardization; history-free endpoints return their original daily representation.

\subsection{Observation-fiber property and proof}
For a sparse observation operator $D$ and a lift $U$ satisfying $DU=I$, let $P=UD$. Then
\begin{equation}
x=Px+(I-P)x,\qquad P^2=P,\qquad D(I-P)x=0.
\label{eq:decomposition}
\end{equation}

Suppose exact whole-view agreement requires $f(x)=g(Dx)$ on all supported signals. For any supported $x+v$ with $Dv=0$, $D(x+v)=Dx$ and consequently $f(x+v)=g(Dx)=f(x)$. The conclusion is constancy within the observation fiber. It concerns the realized residual and exact equality, not whether sparse data predict a residual distribution or whether approximate agreement necessarily destroys empirically useful information. Equation~\ref{eq:decomposition} follows from $P^2=UDUD=UD$ and $D(I-UD)=0$; the lift need not be orthogonal.

\subsection{Clock-aware memory: implementation}
\label{sec:memory-implementation}
The daily encoder and clock-aware memory are components of the single TER encoder. The main text's $z_0$ corresponds to $z_{\rm base}$ below, and $\Delta_\psi=R(\alpha W u_\psi)$ is its memory update.
For fixed scales $\mathcal Q$, the value/change vector in Equation~\ref{eq:binding} is
\[
b_{d,i}=\left[\left(\tanh\frac{g_{d,i}-100}{q}\right)_{q\in\mathcal Q},
\left(\tanh\frac{\Delta g_{d,i}}{q}\right)_{q\in\mathcal Q}\right],
\qquad
\theta_{d,i}=2\pi\,\frac{\operatorname{recordedMinute}_{d,i}}{1440}.
\]
For elapsed day age $a_d$, a learned lift maps
$[s_{d,k},(k/4)s_{d,k},e^{-a_d}s_{d,k},e^{-a_d/3}s_{d,k},e^{-a_d/7}s_{d,k}]$
to tokens. Attention mixes valid day--harmonic tokens: a query of age $a_j$ reads keys of age $a_\ell\geq a_j$, with interaction between same-day harmonics allowed. Residual attention, LayerNorm and a feed-forward block produce $h_{d,k}$, followed by
\begin{equation}
u_\psi=\frac{\sum_{d,k}\nu_d e^{-a_d/3}h_{d,k}}
{\max(10^{-8},\sum_{d,k}\nu_d e^{-a_d/3})},
\qquad z_\psi=z_{\rm base}+R(\alpha W u_\psi),\quad W_0=0 .
\label{eq:writeback}
\end{equation}
Here $\nu_d$ indicates day support and $\alpha$ gates the memory. The earlier context encoder learns the bias-free map $R$ before memory training; $R$ then stays fixed. It is distinct from the episodic Ridge regressor.

Each endpoint uses up to seven nonoverlapping daily windows whose start times span at most six days, within recorded-identity/cohort/split/cadence/source boundaries. The working grid has 288 five-minute positions. Clock-aware memory selects fifteen-minute-compatible positions; fifteen-minute streams use the observed grid residue with greatest support, while the three/five-minute metadata path uses residue zero. The first selected position lacks an adjacent predecessor. Selected adjacent pairs must be finite and observed; padded or missing readings cannot supply innovations. A supported day requires at least 24 such pairs. The memory gate activates only with at least three supported days; shorter histories retain the base representation.

Clock centering at 100 mg/dL, amplitude scales $\mathcal Q=\{10,20,40,80\}$ mg/dL, the hyperbolic tangent, recorded phase, four harmonics, age kernels, and pooling weights are fixed. Phase uses recorded window start plus grid offset. The sine/cosine concatenation in Equation~\ref{eq:binding} is the actual coordinate order. No zero harmonic is used, but a nonuniform mask and nonlinear level basis mean the remaining slots need not be free of average burden.

A bias-free $80\to16$ lift feeds single-head attention with affine Q/K/V and output maps. Invalid/causally disallowed logits receive $-10^9$; softmax weights are multiplied by the allowed mask and renormalized to protect empty rows. A residual, LayerNorm, and $16\to32\to16$ GELU feed-forward network precede fixed recency pooling. The bias-free writeback $W:16\to36$ is zero-initialized; $R:36\to146$ is kept fixed. Clock-aware memory has 4,048 trainable parameters, an implementation detail rather than a performance claim. Its output correction lies in the column space of $R$.

If the clock-aware memory's read/mix parameters are $\eta$ and the representation gradient is $G$, the local terms satisfy
\[
\nabla_W\mathcal L\supset\alpha(R^\top G)u_\eta^\top,\qquad
\nabla_\eta\mathcal L\supset\alpha J_\eta u^\top W^\top R^\top G.
\]
The upstream read/mix loss-gradient is therefore zero at $W=0$; the writeback may receive a nonzero first-step gradient. AdamW weight decay can still affect upstream tensors, so this algebra does not establish a measured two-stage optimization process.

The daily core uses 6,960 AdamW updates, batch size 128, learning rate $3\times10^{-4}$, weight decay $10^{-3}$, and clipping at one. The daily dynamics branch is validation-selected and then fixed. During clock-aware memory training the context encoder stays unchanged. Experiment reports record exact zero error for initialization, inactive-support, and history-free identity checks. Recorded nonzero read/mix and writeback gradients verify that the implemented learning graph is connected; they do not prove the usefulness of each learned component.

The released clock-aware memory uses seed 43, 10,440 AdamW updates, learning rate $4\times10^{-4}$, weight decay $10^{-4}$, and gradient clipping at $0.5$. Only clock-aware memory parameters update, and the last checkpoint is used. Spectral evidence is audit-only and never enters Equation~\ref{eq:outer}. Inference exports one shared representation for all downstream tasks, without an episodic regressor, proxy-target conditioning, or test-time adaptation.

\section{Limited mathematical links to downstream readout}
\label{sec:properties}
For a fixed training set and representation, a sampled episode is a Monte Carlo estimate of the risk induced by the actual sampler in Equation~\ref{eq:generic-objective}. Under differentiability and interchange of expectation and derivative, its unclipped gradient estimates that finite-set objective's gradient. The actual clipped AdamW update is not itself an unbiased gradient estimator. Repeated windows and repeated people must not be counted as independent samples.

Let a continuous downstream target satisfy $y=AH(x)+r(x)$ with $\mathbb E\|r\|^2\le\epsilon^2$. For evidence error $\delta_H=\widehat H-H$ and positive diagonal weights $D_H$,
\[
\mathbb E\|A\widehat H-y\|^2
\le 2\|AD_H^{-1/2}\|_{\mathrm{op}}^2\,
\mathbb E\|D_H^{1/2}\delta_H\|^2+2\epsilon^2 .
\]
Indeed, $A\widehat H-y=A\delta_H-r$; apply $\|u+v\|^2\le2\|u\|^2+2\|v\|^2$ and the operator norm. If $r=0$, the coefficient two is unnecessary. A fixed fit-group-standardized affine evidence regressor composed with $A$ remains affine.

This elementary conditional existence link covers only targets approximated by the bank. It neither guarantees that a newly fitted finite-sample regressor selects this mapping nor bounds PR-AUC or F1. If the bank omits an independent component $v$ of $x=(u,v)$, perfect recovery of $H(x)=u$ says nothing about $y=v$. A shared encoder can memorize signal fingerprints across episodes even with disjoint fit and transfer identity groups. Neither low empirical evidence risk nor grouping alone guarantees invariance or population transfer.

\section{Broad phenotype utility: full benchmark results}
\label{sec:benchmark-details}
Every phenotype uses the same shared representation within an outer episode. Tables~\ref{tab:taskwise} and~\ref{tab:published-taskwise} expand Table~\ref{tab:main}. C, S, St and H denote CGMacros, Shanghai, Stanford and Hall; Lip, Ob, Hypo, Beta and Glu denote hyperlipidemia, obesity, hypoglycemia, beta-cell dysfunction and glucotype. Local checkpoint rows use the same person-grouped folds and linear probe as TER. Published rows retain the source study's reported results.

\paragraph{Best-in-table cells and task coverage.}
A clear best cell exceeds every baseline by more than its rounding interval: .05 percentage points for one-decimal reported values, and zero for unrounded local scores. A task is covered when at least one of its three metrics is a clear best. This yields 30/42 cells across 12/14 tasks. These counts summarize the ten-baseline table; ablation differences use only paired local evaluations.

\begin{table}[H]\centering
\caption{CGM models on all fourteen tasks (\%). Published GlucoFM is from the source paper; other rows use the common local protocol.}
\label{tab:taskwise}
\setlength{\tabcolsep}{2pt}\renewcommand{\arraystretch}{1.13}
\resizebox{\linewidth}{!}{\begin{tabular}{ll*{15}{r}}
\toprule
Model & Metric & C-Dia & C-IR & C-Lip & C-Ob & S-IR & S-Lip & S-Hypo & St-Dia & St-Beta & St-IR & H-Dia & H-IR & H-Lip & H-Glu & Avg\\\midrule
CGMformer & PR & 62.2 & 88.9 & 33.2 & 63.5 & 62.3 & 33.5 & 16.5 & 69.9 & 63.1 & 61.9 & 51.6 & 46.9 & 21.9 & 82.6 & 54.1\\

 & AUC & 76.1 & 79.5 & 56.5 & 64.6 & 50.8 & 50.1 & 56.3 & 67.3 & 63.0 & 63.1 & 60.2 & 54.5 & 48.5 & 88.5 & 62.8\\

 & F1 & 55.0 & 67.4 & 50.8 & 57.6 & 47.5 & 48.0 & 49.2 & 58.4 & 54.8 & 55.2 & 55.5 & 53.3 & 49.9 & 78.9 & 55.8\\

\midrule
CGM-JEPA & PR & 59.2 & 89.7 & 33.6 & 65.8 & 69.1 & 33.6 & 18.5 & 73.5 & 58.5 & 64.3 & 63.4 & 61.3 & 13.6 & 84.8 & 56.3\\

 & AUC & 72.3 & 79.6 & 45.0 & 64.7 & 59.8 & 51.9 & 51.1 & 71.3 & 54.8 & 68.1 & 74.7 & 70.2 & 30.3 & 90.5 & 63.1\\

 & F1 & 48.3 & 66.6 & 48.0 & 58.9 & 53.6 & 41.7 & 46.7 & 59.3 & 47.8 & 60.2 & 64.5 & 60.1 & 45.3 & 80.7 & 55.8\\

\midrule
X-CGM-JEPA & PR & 59.6 & 89.8 & 33.9 & 66.4 & 68.9 & 33.3 & 16.5 & 73.5 & 58.8 & 65.1 & 62.4 & 63.6 & 13.1 & 85.6 & 56.5\\

 & AUC & 73.2 & 79.7 & 46.1 & 65.3 & 59.7 & 49.8 & 49.5 & 71.8 & 55.6 & 69.1 & 74.4 & 71.7 & 28.6 & 90.6 & 63.2\\

 & F1 & 49.9 & 66.6 & 47.5 & 59.2 & 52.7 & 42.0 & 46.8 & 60.1 & 48.2 & 60.9 & 64.7 & 62.6 & 45.4 & 79.9 & 56.2\\

\midrule
GlucoFM (published) & PR & 65.9 & 91.9 & 36.1 & 64.9 & 67.0 & 33.5 & 21.1 & 77.3 & 69.0 & 67.6 & 66.2 & 60.2 & 14.4 & 88.3 & 58.8\\

 & AUC & 78.7 & 81.2 & 54.7 & 62.6 & 57.8 & 50.5 & 59.2 & 72.8 & 68.7 & 69.1 & 75.9 & 70.7 & 41.6 & 90.7 & 66.7\\

 & F1 & 58.3 & 69.6 & 50.2 & 59.4 & 55.4 & 49.1 & 50.7 & 66.2 & 63.3 & 64.0 & 64.5 & 62.0 & 43.1 & 82.4 & 59.9\\

\midrule
TER & PR & 71.4 & 92.4 & 44.3 & 75.7 & 57.1 & 48.4 & 29.6 & 81.4 & 64.5 & 68.8 & 66.6 & 65.3 & 35.1 & 92.0 & 63.8\\

 & AUC & 81.8 & 86.9 & 57.4 & 74.3 & 38.8 & 64.7 & 65.6 & 81.1 & 66.3 & 73.2 & 75.6 & 73.3 & 62.9 & 93.9 & 71.1\\

 & F1 & 60.5 & 78.4 & 54.0 & 63.5 & 36.8 & 54.0 & 50.0 & 67.8 & 56.3 & 60.9 & 64.5 & 64.7 & 50.9 & 85.7 & 60.6\\

\bottomrule\end{tabular}}
\end{table}

\begin{table}[H]\centering
\caption{The generic time-series comparison on all fourteen tasks (\%). Baseline scores are from the GlucoFM study.}
\label{tab:published-taskwise}
\setlength{\tabcolsep}{2pt}\renewcommand{\arraystretch}{1.13}
\resizebox{\linewidth}{!}{\begin{tabular}{ll*{15}{r}}
\toprule
Model & Metric & C-Dia & C-IR & C-Lip & C-Ob & S-IR & S-Lip & S-Hypo & St-Dia & St-Beta & St-IR & H-Dia & H-IR & H-Lip & H-Glu & Avg\\\midrule
Chronos-2 (small) & PR & 51.6 & 82.7 & 30.5 & 56.0 & 65.0 & 34.8 & 16.0 & 69.2 & 58.6 & 62.5 & 51.7 & 44.9 & 24.2 & 54.2 & 50.1\\

 & AUC & 68.0 & 67.5 & 52.1 & 53.7 & 55.0 & 50.6 & 49.1 & 65.4 & 57.3 & 64.2 & 64.9 & 57.7 & 61.3 & 62.6 & 59.2\\

 & F1 & 49.4 & 61.7 & 50.1 & 52.7 & 52.6 & 51.5 & 50.5 & 61.3 & 54.8 & 59.6 & 60.9 & 55.0 & 52.5 & 58.6 & 55.1\\

\midrule
Chronos-2 (120M) & PR & 54.3 & 84.1 & 34.0 & 58.1 & 59.8 & 34.7 & 18.9 & 68.5 & 58.5 & 59.2 & 48.6 & 51.3 & 20.6 & 54.4 & 50.4\\

 & AUC & 69.7 & 69.9 & 55.8 & 56.6 & 49.5 & 49.7 & 53.0 & 65.6 & 57.8 & 60.2 & 60.4 & 62.7 & 50.3 & 63.0 & 58.9\\

 & F1 & 50.2 & 63.3 & 53.7 & 54.2 & 49.0 & 50.4 & 52.8 & 61.9 & 55.5 & 56.6 & 58.4 & 58.2 & 50.9 & 56.2 & 55.1\\

\midrule
MOMENT (small) & PR & 54.0 & 83.8 & 30.3 & 60.9 & 58.3 & 38.3 & 16.3 & 70.4 & 57.2 & 63.0 & 59.2 & 53.1 & 19.4 & 55.6 & 51.4\\

 & AUC & 69.1 & 70.6 & 51.5 & 58.7 & 49.0 & 55.9 & 52.9 & 67.6 & 55.5 & 65.1 & 69.7 & 62.1 & 48.7 & 61.7 & 59.9\\

 & F1 & 49.8 & 63.6 & 50.9 & 55.7 & 49.3 & 52.9 & 51.4 & 62.5 & 53.8 & 60.1 & 64.5 & 59.0 & 50.4 & 59.3 & 55.9\\

\midrule
MOMENT (large) & PR & 50.3 & 87.4 & 32.0 & 61.8 & 59.6 & 42.4 & 14.3 & 71.9 & 61.8 & 58.1 & 52.1 & 44.9 & 18.4 & 55.8 & 50.8\\

 & AUC & 66.1 & 75.2 & 53.2 & 59.9 & 47.8 & 60.8 & 49.6 & 69.1 & 58.9 & 59.4 & 65.0 & 56.6 & 51.4 & 63.2 & 59.7\\

 & F1 & 46.8 & 67.2 & 52.5 & 56.3 & 48.6 & 57.2 & 49.7 & 63.7 & 55.7 & 56.1 & 60.5 & 53.3 & 47.7 & 57.8 & 55.2\\

\midrule
Mantis & PR & 61.8 & 90.9 & 30.1 & 63.4 & 59.2 & 30.6 & 22.9 & 75.1 & 63.5 & 67.1 & 54.2 & 55.7 & 24.8 & 77.5 & 55.5\\

 & AUC & 75.5 & 80.3 & 49.8 & 60.8 & 48.2 & 46.0 & 58.6 & 71.4 & 61.5 & 68.2 & 67.1 & 66.2 & 57.9 & 82.6 & 63.9\\

 & F1 & 56.0 & 71.6 & 49.8 & 57.2 & 48.4 & 47.5 & 53.8 & 65.9 & 58.5 & 62.3 & 59.6 & 59.8 & 53.7 & 73.3 & 58.4\\

\midrule
MantisV2 & PR & 63.6 & 91.2 & 30.3 & 62.8 & 64.1 & 33.3 & 22.0 & 75.2 & 65.7 & 64.6 & 57.7 & 59.3 & 25.0 & 81.5 & 56.9\\

 & AUC & 76.7 & 80.3 & 49.1 & 60.3 & 52.7 & 48.5 & 59.6 & 71.2 & 64.4 & 65.5 & 67.5 & 66.7 & 58.8 & 84.7 & 64.7\\

 & F1 & 56.2 & 68.9 & 49.2 & 55.9 & 51.2 & 48.1 & 53.3 & 65.7 & 58.6 & 60.4 & 61.1 & 60.8 & 52.6 & 76.7 & 58.5\\

\midrule
GlucoFM (published) & PR & 65.9 & 91.9 & 36.1 & 64.9 & 67.0 & 33.5 & 21.1 & 77.3 & 69.0 & 67.6 & 66.2 & 60.2 & 14.4 & 88.3 & 58.8\\

 & AUC & 78.7 & 81.2 & 54.7 & 62.6 & 57.8 & 50.5 & 59.2 & 72.8 & 68.7 & 69.1 & 75.9 & 70.7 & 41.6 & 90.7 & 66.7\\

 & F1 & 58.3 & 69.6 & 50.2 & 59.4 & 55.4 & 49.1 & 50.7 & 66.2 & 63.3 & 64.0 & 64.5 & 62.0 & 43.1 & 82.4 & 59.9\\

\midrule
TER & PR & 71.4 & 92.4 & 44.3 & 75.7 & 57.1 & 48.4 & 29.6 & 81.4 & 64.5 & 68.8 & 66.6 & 65.3 & 35.1 & 92.0 & 63.8\\

 & AUC & 81.8 & 86.9 & 57.4 & 74.3 & 38.8 & 64.7 & 65.6 & 81.1 & 66.3 & 73.2 & 75.6 & 73.3 & 62.9 & 93.9 & 71.1\\

 & F1 & 60.5 & 78.4 & 54.0 & 63.5 & 36.8 & 54.0 & 50.0 & 67.8 & 56.3 & 60.9 & 64.5 & 64.7 & 50.9 & 85.7 & 60.6\\

\bottomrule\end{tabular}}
\end{table}

\paragraph{Public-only GlucoFM reproduction.}
The matched public-only row uses our independently initialized GlucoFM implementation on the same 2,283 unlabeled windows as public-only TER (1,868 for fitting and 415 for label-free validation). It uses 24 one-hour patches per daily window, a 128-channel, three-layer/four-head context encoder, a one-layer predictor, and an EMA target. Equal-weight masked-context and state/event transition losses mask 50--60\% of patches; the optional sparse-aware masking and TER components are off. Training uses seed 43, 120 epochs, batch size 128, relative auxiliary features, and the historical reproduction's cadence/cohort sampling settings (15-minute share .15; cohort temperature .5). Its checkpoint is selected by unlabeled validation loss; the 14 phenotype tasks use the same subject-grouped folds and StandardScaler--logistic probe as TER. This comparison matches pretraining records and downstream evaluation, not architecture or optimization budget; the official GlucoFM code and checkpoint were unavailable.

\section{Clock-aware memory: component controls}
\label{sec:components}\label{sec:full-budget-detail}
These fixed-configuration component interventions use the same trained multi-day base, final daily/dynamics inputs, and 6,960 memory-stage updates. Each control changes one component and is evaluated at its terminal checkpoint, on the same outer folds as the main table. The reference is the matching fixed-configuration checkpoint, rather than the main table's inner-selected 24-candidate procedure. No encoder is refitted for downstream evaluation.

\begin{table}[H]\centering\small
\caption{Fixed-configuration clock components on the common outer folds (\%).}
\label{tab:full-budget-absolute}
\setlength{\tabcolsep}{4pt}
\begin{tabular*}{\linewidth}{@{\extracolsep{\fill}}lrrr@{}}
\toprule
Representation & PR-AUC & ROC-AUC & Macro-F1\\\midrule
Full matched reference & 64.53 & 71.81 & 60.96\\
Reassigned value--time rows & 62.89 & 70.58 & 59.89\\
No day age/recency & 64.10 & 71.50 & 60.61\\
No internal causal mask & 64.56 & 71.75 & 60.88\\
No harmonic 1 & 63.45 & 71.07 & 60.54\\
No harmonic 2 & 64.25 & 71.88 & 60.45\\
No harmonic 3 & 64.29 & 71.51 & 60.91\\
No harmonic 4 & 63.79 & 70.84 & 60.61\\
\bottomrule\end{tabular*}
\end{table}

\begin{table}[H]\centering\small
\caption{Matched reference minus each removal (percentage points). Effects are conditional, not additive.}
\label{tab:components}
\setlength{\tabcolsep}{4pt}
\begin{tabular*}{\linewidth}{@{\extracolsep{\fill}}lrrr@{}}
\toprule
Removal & $\Delta$ PR & $\Delta$ ROC & $\Delta$ F1\\\midrule
Reassigned value--time rows & 1.64 & 1.23 & 1.06\\
No day age/recency & 0.43 & 0.31 & 0.34\\
No internal causal mask & -0.02 & 0.06 & 0.08\\
No harmonic 1 & 1.08 & 0.74 & 0.42\\
No harmonic 2 & 0.28 & -0.06 & 0.50\\
No harmonic 3 & 0.24 & 0.30 & 0.04\\
No harmonic 4 & 0.74 & 0.97 & 0.34\\
\bottomrule\end{tabular*}
\end{table}

\paragraph{Exact interventions.}
The day-age ablation zeros all three explicit age-interaction blocks and replaces exponential recency weighting with uniform valid-token pooling. Chronological ranks remain only for the causal mask. The bidirectional ablation removes that mask while retaining the same strictly pre-endpoint inventory, age features, and recency readout. Each harmonic ablation zeros one entire sine/cosine block. Nominal parameter counts remain the same even when an input block is zero; this is an input/component intervention, not a claim that every parameter retains identical effective capacity.

\paragraph{Matched clock assignment.}
The separate clock control applies a fixed even-index-then-odd-index permutation to complete active numerical rows before the unchanged projection. It preserves the values, masks, day support, clock basis, and capacity, but changes their physical assignment and within-day numerical order. Full-budget dedicated EV is .1227826755 for the true clock and .0859481515 for the permuted control (difference .0368345240). Their event EVs are .4757675976 and .4741128907; episode losses are 1.3014593124 and 1.2813862562. These auxiliary readouts are diagnostic, not phenotype-selection criteria.

All seven interventions are reported above. The causal-mask control retains only pre-endpoint observations even when internal attention is bidirectional. Training-only spectral recovery and phenotype utility measure different outcomes.

\section{Observation-aware daily encoding: paired core control}
\label{sec:signed-core-control}
For Observation-aware daily encoding, this historical core-stage experiment isolates signed dense-only targets, not a removal from the final system. The two cores use the same exact known-cadence projector, relative factor-three observation construction, data, seed 43, capacity, and 6,960-update budget. Only the unsigned versus upper/lower signed target changes. Neither core includes the final multi-day context encoder or clock-aware memory stage.

\begin{table}[H]\centering
\caption{Observation-aware daily encoding: a matched historical core comparison. Higher phenotype means are better; lower dense-only q95 tail-profile NMSE is better. These are core endpoints, not alternative final models. NMSE diagnoses this tail target only; it is not comparable to the spectral EV diagnostics.}
\label{tab:signed-core-control}
\small\setlength{\tabcolsep}{3pt}\renewcommand{\arraystretch}{1.15}
\begin{tabular*}{\linewidth}{@{\extracolsep{\fill}}lrrrr@{}}
\toprule
Core targets & PR-AUC & ROC-AUC & Macro-F1 & q95 NMSE\\
\midrule
Unsigned dense-only & .519523 & .608946 & .561858 & .968126\\
Signed upper/lower & .528276 & .615488 & .565317 & .794457\\
\bottomrule\end{tabular*}
\end{table}

The signed-minus-unsigned phenotype changes are +.875/.654/.346 percentage points, with 26 improved and 16 decreased task--metric cells. Dense-only q95 NMSE falls by 17.94\%, with upper/lower reductions of 36.13\% and 3.83\%; q95 identifies the tail-profile target, not an error quantile. Common-view cosine distances are numerical zero in both variants. This supports signed evidence targets in observation-aware daily encoding, not independent projector necessity, final-system ablation gains, or additive attribution across training stages.

\section{Paired downstream improvement counts}\label{sec:paired-cells}
\begin{table}[H]\centering\small
\caption{Strict improvements across the 42 paired task--metric means on common outer folds.}
\label{tab:paired-cells}
\setlength{\tabcolsep}{4pt}
\begin{tabular*}{\linewidth}{@{\extracolsep{\fill}}lrrr@{}}
\toprule
Comparison & Improved & Tied & Decreased\\\midrule
TER vs direct evidence & 35 & 0 & 7\\
TER vs daily + evidence & 35 & 0 & 7\\
Matched TER vs mean--max & 38 & 0 & 4\\
Matched TER vs BiLSTM & 39 & 0 & 3\\
\bottomrule\end{tabular*}
\end{table}

\section{Pretraining corpus and downstream dataset inventory}
\label{sec:data-inventory}
\paragraph{Pretraining combines private and public unlabeled CGM.}
Across the pretraining corpora, the union contains 23,336 private windows by window ID and the same 2,283 public windows. The private source contributes only de-identified, unlabeled glucose, observation masks, and timing/source metadata, not phenotype annotations; the records were used with user authorization for research. Raw or person-level private data are not distributed. Table~\ref{tab:pretraining-inventory} reports the stage-level allocations needed to reproduce training; they are not additive or nested, and per-source biological-participant counts are not established by the aggregate manifests.

\begin{table}[H]\centering
\caption{Stage-specific unlabeled pretraining windows. Private rows are separately selected across stages; counts are neither additive nor nested. The fixed 2,283-window public component comprises BIG IDEAs (211), Col\'as (490), ShanghaiT2DM (948), and Stanford (634).}
\label{tab:pretraining-inventory}
\small\setlength{\tabcolsep}{5pt}
\begin{tabular*}{\linewidth}{@{\extracolsep{\fill}}lrrrl@{}}
\toprule
Stage & Private & Public & Total & Role\\\midrule
Context and memory base & 6,138 & 2,283 & 8,421 & Shared longitudinal base\\
Daily representation & 7,673 & 2,283 & 9,956 & Anchor-preserving extension\\
Dynamics continuation & 10,742 & 2,283 & 13,025 & Evidence-balanced 1.75$\times$ coreset\\
\bottomrule\end{tabular*}
\end{table}

\paragraph{Fitting, audit, and exclusion.}
The base corpus stores 7,341 fitting and 1,080 audit windows and excludes 649 private-source held-out windows. The daily extension has 8,876 fitting and 1,080 audit windows; the final dynamics allocation has 11,446 fitting and 1,579 validation windows. Optimizer updates use only fitting rows. The base support audit finds 3,156 endpoints with at least two nonoverlapping days and 1,440 with at least three; these counts precede the memory's additional observed-pair support gate. History never crosses stored identity, cohort, split, cadence, or source, and spans at most seven days. Shanghai and Stanford occur in both source collections and the benchmark, but the audited build excludes their identified downstream participants. Appendix~\ref{sec:identity-scope} specifies the identity linkage and evaluation protocol.

\paragraph{Final local-branch recipe.}
``Anchor-preserving'' denotes a data and exposure policy, not an additional loss: the daily allocation retains every row of a separately quality-screened 8,421-window anchor, adds 1,535 fitting rows, and scales the update count to preserve expected anchor-row exposure. The daily encoder is trained from scratch with seed 43 for 38,527 AdamW updates (batch 128, learning rate $3\times10^{-4}$, weight decay $10^{-3}$, gradient clipping at one), using the unchanged cadence/source sampler and threefold rare-event weight; its terminal checkpoint is retained. For the dynamics continuation, candidate private rows pass the fixed glucose/missingness quality screen and exclude exact base duplicates. We compute the four 16-coordinate evidence families of Appendix~\ref{sec:target-details}, standardize them using public evidence only, clip to $[-8,8]$, and fit 32-means clustering to the public rows (seed 43, ten initializations). Split/source quotas preserve the base mixture; within each quota, rows nearest public cluster centroids are selected, with one row per stored identity when possible. The resulting 13,025-window allocation initializes the fixed preceding dynamics checkpoint and runs 15,944 AdamW updates with the same batch size, learning rate, weight decay, architecture, and label-free objective; the checkpoint with minimum validation total at the fixed reporting intervals is retained. Finally, the affine physical-scale coordinates are recomputed from the adapted daily checkpoint's saved center and scale. Assembly retains the fixed context and memory checkpoints, substitutes the daily, dynamics, and physical-scale components, verifies their saved corpus/configuration hashes, and re-extracts all downstream representations. These within-run checkpoints use the terminal-update or unlabeled-validation rules specified above; cross-run model selection is described in Appendix~\ref{sec:model-selection}.

\paragraph{Downstream evaluation units.}
Table~\ref{tab:downstream-inventory} reports the processed participants and windows scored, not raw-study enrollment. All tasks within a cohort share the same biological-person-grouped partitions. Ten repeats of five outer folds and three training-internal folds follow Appendix~\ref{sec:model-selection}; all visits of a person remain together. Each training fold fits a StandardScaler and scikit-learn logistic regression ($C=1$, L2 penalty, L-BFGS, 1,000 maximum iterations, no class weighting); the resulting model scores only held-out rows. These labels do not enter representation learning. PR-AUC uses average precision; multiclass PR-AUC and ROC-AUC use macro one-versus-rest averaging, and Macro-F1 uses the highest-probability class. Metrics are averaged over folds/repeats, and Overall weights the 14 tasks equally.

\begin{table}[H]\centering
\caption{Processed downstream cohorts. Cadence is in minutes; biological participant is the split unit. Shanghai's 65 visits are grouped into 58 participants.}
\label{tab:downstream-inventory}
\small\setlength{\tabcolsep}{4pt}
\begin{tabular*}{\linewidth}{@{\extracolsep{\fill}}lrrrr@{}}
\toprule
Cohort & Cadence & Participants & Windows & Tasks\\\midrule
CGMacros & 5/15 & 45 & 855 & 4\\
ShanghaiT2DM & 15 & 58 & 609 & 3\\
Stanford & 5 & 37 & 994 & 3\\
Hall & 5 & 56 & 250 & 4\\
\bottomrule\end{tabular*}
\end{table}

CGMacros covers diabetes, hyperlipidemia, insulin resistance, and obesity; ShanghaiT2DM covers insulin resistance, hyperlipidemia, and hypoglycemia; Stanford covers diabetes, insulin resistance, and beta-cell dysfunction; Hall covers diabetes, insulin resistance, hyperlipidemia, and glucotype. The common-protocol CGM comparisons use these identical folds and probe settings. Published generic-model results retain their separate source-study provenance. The GlucoFM paper-reported row likewise retains its original protocol and a pretraining corpus containing non-public Wear-CGM data; because neither that corpus nor its trained checkpoint is released, it is explicitly labeled as a reported rather than reproduced value in Table~\ref{tab:main}.

\clearpage
\section{Independent AIDE temporal and cohort validation}
\label{sec:aide-external}

\paragraph{Seven-outcome grouped-fold evaluation.}
For the longitudinal aggregation comparison, we use 81 eligible participants and
236 treatment periods with five participant-grouped folds. Of the 238 periods
with eligible input histories, two lack the required future-CGM coverage for
the seven outcomes. Each model
receives the same 4--7 chronological daily representations from Weeks 5--6 and the same
treatment covariate; a fold-local StandardScaler and Ridge regressor
($\alpha=1$) predict seven future CGM outcomes from Weeks 7--12. The BiLSTM
is the previously trained, label-free public-benchmark comparator, applied
without any AIDE outcome training of its representation. The seven outcomes
and their participant-weighted mean absolute errors are in Table~\ref{tab:aide-seven-outcomes}.

\begin{table}[H]\centering
\caption{\textbf{AIDE longitudinal transfer.} Participant-weighted MAE for seven future CGM outcomes (lower is better); daily inputs, folds, and treatment covariate are matched.}
\label{tab:aide-seven-outcomes}
\footnotesize\setlength{\tabcolsep}{4pt}\renewcommand{\arraystretch}{1.0}
\begin{tabular*}{\linewidth}{@{\extracolsep{\fill}}lrrr@{}}
\toprule
Future outcome & Mean--max & BiLSTM & TER\\\midrule
Time $<70$ mg/dL (\%) & 2.172 & 1.879 & \textbf{1.143}\\
Time $<54$ mg/dL (\%) & .739 & .539 & \textbf{.395}\\
$<54$ mg/dL events/week & 1.518 & 1.307 & \textbf{.858}\\
Time 70--180 mg/dL (\%) & 12.667 & 13.468 & \textbf{6.118}\\
Time $>180$ mg/dL (\%) & 13.134 & 13.757 & \textbf{6.123}\\
Mean glucose (mg/dL) & 20.463 & 22.059 & \textbf{9.738}\\
Glucose CV (\%) & 6.320 & 6.256 & \textbf{3.146}\\
\bottomrule
\end{tabular*}
\end{table}

\paragraph{Separate held-out hypoglycemia evaluation.}
The public AIDE T1D study is a multicenter randomized three-period crossover trial of hybrid closed loop, predictive low-glucose suspend, and sensor-augmented pump therapy in 82 adults aged at least 65 years \citep{kudva2025aide}. It is absent from TER pretraining and from candidate selection. Before reading held-out outcomes, we assign 48/18/16 participant identities to discovery/validation/private test from the randomized sequence list; every treatment period from one participant stays in one split.

For each eligible treatment period, the fixed input is the earliest seven-calendar-day span in Weeks 5--6. We emit only its 4--7 days containing at least eight physical CGM readings and never interpolate a fully unobserved day. This yields 238 periods and 1,656 daily windows. The physical-statistics comparator contains 21 prespecified summaries of all observed input values: mean, standard deviation, coefficient of variation, minimum, maximum, nine quantiles (1/5/10/25/50/75/90/95/99\%), fractions below 54 and 70, within 70--180, above 180 and 250 mg/dL, and the mean absolute and standard deviation of adjacent observed changes. The later endpoint is percentage of CGM readings below 70 mg/dL in non-overlapping Weeks 7--12. A discovery-only StandardScaler and Ridge regressor ($\alpha=1$) is fitted once for each fixed input, always with the same HCL/PLGS/SAP one-hot covariate. Validation is opened once without tuning, followed by one opening of the private test.

\begin{table}[H]\centering
\caption{Independent AIDE later-hypoglycemia prediction. MAE is participant weighted; higher Spearman is better. The private test contains 15 eligible participants and 44 periods.}
\label{tab:aide-external-full}
\small\setlength{\tabcolsep}{5pt}\renewcommand{\arraystretch}{1.08}
\begin{tabular*}{\linewidth}{@{\extracolsep{\fill}}lrrrr@{}}
\toprule
& \multicolumn{2}{c}{Validation} & \multicolumn{2}{c}{Private test}\\
Fixed input & MAE & $\rho$ & MAE & $\rho$\\\midrule
Daily representations, mean--max & 2.220 & $-.033$ & 1.833 & .150\\
Seven-day physical statistics & 1.227 & .489 & \textbf{.962} & .621\\
\textbf{TER} & \textbf{1.195} & \textbf{.560} & .969 & \textbf{.631}\\
\bottomrule
\end{tabular*}
\end{table}

On the private test, TER-minus-mean--max MAE is $-.865$ with paired-participant bootstrap 95\% interval $[-1.381,-.329]$. TER-minus-physical-statistics MAE is $+.006$, interval $[-.179,+.180]$. The representation archive was fixed before endpoint construction and contains no endpoint columns; the discovery, validation, and private-test identity intersections are empty. The result supports transfer to unseen participants and a later period, with a large gain over conventional mean--max pooling. The interval against the endpoint-aligned physical summary includes zero; this does not establish either a difference or statistical equivalence, treatment-effect estimation, or superiority to clinical features.

\clearpage
\section{Postprandial response transfer}
\label{sec:ppgr-transfer}

\paragraph{Cohort construction and prediction protocol.}
The linkage begins with 2,132 logged meals from 469 people. Excluding meals followed by another logged meal within two hours leaves 1,979 candidates. Requiring one strictly causal 24-hour CGM history and physical glucose measurements at all 24 post-meal target times (5--120 minutes) leaves 1,705 meals from 409 people. Median pre-meal history coverage is 100\% (minimum 96.18\%); the median and maximum baseline lag are 1.70 and 4.98 minutes, and the median and maximum event-level target-alignment errors are 1.28 and 2.50 minutes. The same device source supplies each event's history and outcome. With only one day of history, TER's multi-day context and clock memory are inactive; this benchmark evaluates the shared daily encoder rather than the longitudinal stages.

All five representations are held fixed. The representation-only predictor receives only the 24-hour embedding. The complete predictor adds the same last hour of pre-meal CGM and current-meal calories, carbohydrate, protein, fat, and food count. Five fixed subject-disjoint folds and ten head initializations are shared across models. Within every training fold, inputs and the 24 physical glucose-change targets are standardized before fitting the same 128/64 ReLU MLP for 600 full-batch epochs. We report subject-macro MAE; lower is better.

\begin{table}[H]\centering
\caption{\textbf{Postprandial response prediction on 1,705 meals from 409 people.} Values are subject-macro MAE averaged over ten head initializations. Panel A isolates representation transfer; Panel B supplies every representation with identical recent-CGM and meal context. Bold denotes the best mean within each panel. Trajectory and peak rise are in mg/dL, positive iAUC in mg/dL$\cdot$h, and peak time in minutes.}
\label{tab:ppgr-full}
\scriptsize\setlength{\tabcolsep}{3.5pt}\renewcommand{\arraystretch}{1.10}
\begin{tabular*}{\linewidth}{@{\extracolsep{\fill}}lrrrr@{}}
\toprule
Representation & Trajectory & Pos. iAUC & Peak rise & Peak time\\
\midrule
\multicolumn{5}{l}{\textit{A. Fixed representation only}}\\
CGMformer & 23.909 & 26.834 & 25.720 & 40.367\\
CGM-JEPA & 21.811 & 24.899 & 24.956 & 34.770\\
X-CGM-JEPA & 21.218 & 23.607 & 23.814 & \textbf{34.490}\\
GlucoFM reimplementation & 22.254 & 25.599 & 24.709 & 36.314\\
\textbf{TER} & \textbf{20.220} & \textbf{22.496} & \textbf{22.857} & 35.803\\
\midrule
\multicolumn{5}{l}{\textit{B. Fixed representation + shared recent-CGM and meal context}}\\
CGMformer & 20.510 & 23.328 & 23.252 & 35.898\\
CGM-JEPA & 19.037 & 22.092 & 22.281 & 31.433\\
\textbf{X-CGM-JEPA} & \textbf{18.748} & \textbf{21.563} & \textbf{21.918} & \textbf{31.103}\\
GlucoFM reimplementation & 20.685 & 24.938 & 24.278 & 33.263\\
TER & 19.297 & 21.861 & 22.084 & 33.664\\
\bottomrule
\end{tabular*}
\end{table}

\paragraph{Representation-only transfer.}
TER is best on trajectory, positive iAUC, and peak rise, reducing error by 4.70\%, 4.71\%, and 4.02\% relative to the strongest non-TER mean. X-CGM-JEPA is best on peak time, where TER is 3.81\% higher. Paired subject bootstraps place TER's trajectory difference from X-CGM-JEPA below zero; the favorable iAUC and peak-rise intervals cross zero. TER is significantly better than CGM-JEPA and the GlucoFM reimplementation on the first three endpoints.

\paragraph{Complete predictor and identity sensitivity.}
With shared recent-CGM and meal context, X-CGM-JEPA is best on all four endpoints. TER nevertheless reduces GlucoFM's trajectory, positive-iAUC, and peak-rise errors by 6.71\%, 12.34\%, and 9.04\%, with paired subject-bootstrap intervals below zero; peak-time error is 1.21\% higher and indistinguishable. The powered cohort may include identities that contributed unlabeled CGM to TER pretraining, although no postprandial outcome, meal label, or downstream fold enters representation training or model selection. For the complete predictor, a stricter sensitivity analysis removes every such identity, leaving 130 meals from 39 people. TER's mean errors are lower than GlucoFM's on all four endpoints by 3.15\%, 3.76\%, 6.00\%, and 15.55\%, respectively, but all intervals cross zero because this subset is small. We therefore use the full cohort for power and the smaller split only as an identity-disjoint directional check; neither is an external clinical validation.

\clearpage
\section{Evidence-recovery audits}
\label{sec:reader-lifecycle-audits}

\paragraph{Full-pipeline control integrity.}
The matched joint-decoder and episodic-Ridge arms complete 6,960 context-stage and 10,440 memory-stage updates. Their corpus, daily input, evidence targets, seed, output dimension, and outer folds match. Joint linear decoders are Ridge-initialized and optimized with the encoder throughout training; both arms use terminal checkpoints and eight candidates selected by the same inner-CV rule. Representation learning uses signal-derived targets, not phenotype labels.

\paragraph{Paired task--metric result.}
The paired comparison subtracts the joint-decoder arm's unrounded score from the matched episodic-Ridge arm for each of the 14 tasks and three metrics. Episodic recovery is higher in 39 of 42 cells and in at least two metrics for 13 of 14 tasks; Shanghai insulin resistance is the exception. This compares recovery procedures, not SOTA coverage against external baselines.

\paragraph{Interpretation.}
The matched $2\times2$ comparison in Appendix~\ref{sec:objective-isolation} separates target choice from the recovery procedure. The same-solver intervention in Appendix~\ref{sec:fit-transfer-intervention} then isolates cross-group scoring. Together they connect the phenotype gains to the learning objective rather than merely supplying additional evidence as downstream features.

\section{Long-context alternatives on common outer folds}
\label{sec:label-history-transfer}
All models are scored on identical endpoints and the primary person-grouped outer folds. Mean--max pools the same observed daily representations; BiLSTM learns reconstruction from these daily representations with a one-layer bidirectional encoder. The matched TER reference uses the same daily inputs and histories. The raw seven-day encoder instead extends local raw-signal processing to seven days and trains for 38,527 updates. No phenotype label trains any representation.

\begin{table}[H]\centering\small
\caption{Long-context alternatives (\%). Matched component reference is used for the pooling contrasts; the final row repeats the main-table evaluation.}
\label{tab:fixed-history-pooling}
\setlength{\tabcolsep}{4pt}
\begin{tabular*}{\linewidth}{@{\extracolsep{\fill}}lrrr@{}}
\toprule
Representation & PR-AUC & ROC-AUC & Macro-F1\\\midrule
Daily only & 54.89 & 64.48 & 56.68\\
Mean--max history pooling & 55.86 & 65.92 & 57.34\\
BiLSTM history pooling & 55.30 & 63.66 & 55.93\\
TER, matched component reference & 64.53 & 71.81 & 60.96\\
Seven-day raw encoder & 57.46 & 66.07 & 57.55\\
TER, main-table selection & 63.75 & 71.13 & 60.56\\
\bottomrule\end{tabular*}
\end{table}

The matched TER reference exceeds mean--max by 8.67/5.89/3.62 points and BiLSTM by 9.24/8.16/5.03 points. The BiLSTM's label-free reconstruction loss falls from .2617 to .1441 during training. Main-table TER exceeds the raw seven-day encoder by 6.29/5.06/3.01 points.

\section{Task definitions and their relation to signal evidence}
\label{sec:task-evidence-map}
Table~\ref{tab:task-evidence-map} orders endpoints by their definitional connection to CGM, not a measured correlation or an empirical ranking of task difficulty. The four groups are: \textbf{A}, CGM-derived pattern or recorded event (two tasks); \textbf{B}, laboratory/diagnostic glycemic status (three); \textbf{C}, physiological proxies (five); and \textbf{D}, more indirect anthropometric/lipid correlates (four). Group A deliberately distinguishes Hall's CGM-derived glucotype from Shanghai's separately supplied visit-level hypoglycemia annotation. Neither is obtained by copying our evidence targets into the downstream label.

\begin{table}[H]\centering
\caption{All 14 benchmark task definitions, checked against the fixed protocol. Evidence examples are qualitative motivation, not task-specific targets or measured associations. HOMA-IR and SSPG are insulin-resistance measures; DI is disposition index. Lipid cutoffs are total cholesterol $\ge240$, LDL $\ge160$, or triglycerides $\ge200$ mg/dL, after source-unit conversion.}
\label{tab:task-evidence-map}
\fontsize{8.5}{10}\selectfont\setlength{\tabcolsep}{3pt}\renewcommand{\arraystretch}{1.12}
\begin{tabular*}{\linewidth}{@{\extracolsep{\fill}}>{\raggedright\arraybackslash}p{.25\linewidth}>{\raggedright\arraybackslash}p{.39\linewidth}>{\raggedright\arraybackslash}p{.29\linewidth}@{}}
\toprule
Task (group) & Source of benchmark label & Plausible signal evidence\\\midrule
Hall glucotype (A) & Provided CGM-pattern class $=2$ & Distribution, variability, excursions\\
Shanghai hypoglycemia (A) & Visit-summary yes/no annotation & Low burden, tails, duration\\\midrule
CGMacros diabetes (B) & HbA1c strata: $<5.7$, $[5.7,6.5)$, $\ge6.5$ & Level, upper tail, burden\\
Stanford diabetes (B) & HbA1c $\ge5.7$ & Level, upper tail, burden\\
Hall diabetes (B) & Provided diagnosis code $>0$ & Level, upper tail, burden\\\midrule
CGMacros insulin resistance (C) & Fasting HOMA-IR $>2.9$ & Rise/recovery, burden\\
Shanghai insulin resistance (C) & Fasting HOMA-IR $>2.9$ & Rise/recovery at observed cadence\\
Stanford insulin resistance (C) & Released SSPG-based IR class & Repeated rise/recovery\\
Hall insulin resistance (C) & SSPG $>120$ when available; otherwise HOMA-IR $>2.9$ & Rise/recovery, distribution\\
Stanford beta-cell dysfunction (C) & Released DI median-split class & Rise/recovery, event variability\\\midrule
CGMacros obesity (D) & BMI $\ge30$ & Possible distribution/pattern correlates\\
CGMacros hyperlipidemia (D) & Laboratory lipid cutoffs & Possible distribution/variability correlates\\
Shanghai hyperlipidemia (D) & Laboratory lipid cutoffs & Possible distribution/variability correlates\\
Hall hyperlipidemia (D) & Laboratory lipid cutoffs & Possible distribution/variability correlates\\
\bottomrule\end{tabular*}
\end{table}

The published datasets separate CGM measurements from laboratory, anthropometric, and clinical endpoints \citep{das2025cgmacros,zhao2023chinese,metwally2025subphenotypes,hall2018glucotypes}. The benchmark's ``diabetes'' names are retained, but some mappings include prediabetic glycemia; they must not be read as uniformly diagnosed diabetes. Glucotype originates in CGM-pattern clustering, while SSPG and DI derive from metabolic tests. The label-definition audit reads protocol code and saved summaries, not private phenotype arrays.

CGM burden and time-in-range summaries have established interpretive meaning \citep{battelino2019tir}; free-living rise/recovery remains an indirect physiological proxy, confounded by unannotated meals, activity, and acquisition. Obesity and lipids are still further removed: CGM need not identify them. These relations motivate a common set of evidence targets $H$, not a promise that it is sufficient for every endpoint. Stable level/distribution information is not necessarily sparse. The schematic duration-versus-evidence argument is not a measured failure of contrastive learning or MAE, and no task-specific branch, evidence subset, or regressor is selected during pretraining.

\clearpage
\section{Learning lineage and limits of interpretation}
\label{sec:prior-scope}
\paragraph{Shared SSL and episodic-learning ideas.}
Paired-view contrastive learning and masked reconstruction both construct label-free supervision. They are not restricted to images. A semantic image label normally requires annotations rather than a prespecified pixel summary; medically meaningful signal summaries offer another possible target, not a universal replacement. Transferable evidence reconstruction connects evidence recovery with stability of the decoding rule; it is neither a sum of losses nor an empirical ranking of SSL families.

Bilevel representations \citep{franceschi2018bilevel}, R2-D2's differentiable Ridge regression \citep{bertinetto2019r2d2}, and MetaOptNet's convex learners \citep{lee2019metaoptnet} establish the optimization lineage. CACTUs, STUNT, and UMTRA construct unlabeled episodes \citep{hsu2019cactus,nam2023stunt,khodadadeh2019umtra}. Our formulation specifies continuous signal evidence and an episodic fit--transfer objective for repeated observations; it does not introduce bilevel or unsupervised meta-learning.

\paragraph{Signal-derived supervision and temporal representations.}
Wearable-derived features already supply self-supervised targets \citep{merrill2023sensing}, and SelfReplay uses sensory meta-SSL \citep{yoon2025selfreplay}. DeepTime combines Fourier representations and Ridge for within-series forecasting \citep{woo2023deeptime}, with CGM time-index meta-optimization in MOTIM \citep{cui2024motim}. GlucoFM uses state/event latent objectives, CGMformer masks glucose tokens, and CGM-JEPA predicts masked latents; X-JEPA adds glucodensity \citep{li2026glucofm,lu2025cgmformer,muhammad2026cgmjepa}. Our evaluated instance trains the observation-aware and clock-aware CGM representation through transferable evidence reconstruction; descriptors and temporal encoders are not claimed as new.

\paragraph{Scope of the controlled comparisons.}
Daily observation-aware encoding has a matched historical core test; clock-aware memory has matched final-model controls. Transferable evidence reconstruction has a complete structured-versus-raw target $\times$ joint-decoder-versus-episodic-Ridge-regression ablation, plus direct-evidence and same-solver cross-group-scoring controls under the common public14 protocol (Appendix~\ref{sec:objective-isolation}). These controls characterize the tested TER training procedure rather than establishing a universal SSL ranking; broad contrastive and masked-autoencoding families are not exhaustively compared. Low evidence error alone guarantees neither arbitrary phenotype utility nor successful finite-sample Ridge regression (Appendix~\ref{sec:properties}).

\paragraph{Identity, exposure, and clinical scope.}
The loss asks a rule to cross chosen groups, not to erase identity or acquisition cues. Stored Shanghai identities are visits rather than globally resolved biological participants; cross-source linkage remains incomplete. Appendix~\ref{sec:model-selection} reports the cross-validation selection protocol and candidate-selection rule. AIDE evaluates disjoint participants and future periods in older adults with type 1 diabetes. Together, these studies cover phenotype prediction and future glucose outcomes; their scope is retrospective CGM representation learning, not real-time diagnosis or treatment benefit. Extension to other medical or wearable signals remains to be tested.

\clearpage
\section{Multistage training procedure}
\label{sec:multistage-contract}
\begin{algorithm}[t]
\caption{Complete training and final-assembly procedure}
\label{alg:multistage}
\fontsize{7.5}{8.3}\selectfont\setlength{\fboxsep}{2pt}
\begin{algorithmic}[1]
\Require Unlabeled pool $\mathcal P$; evidence $H$; auxiliary pool/evidence $\mathcal P_o,H_o$; weight $\lambda_o$ (Appendix~\ref{sec:model-selection}).
\algstage{A}{Learn the base daily representations}
\State Train the observation-aware daily encoder and an auxiliary dynamics encoder with their label-free objectives.
\State Fix both base encoders; append fixed affine features and pack each endpoint's observed history.
\algstage{B}{Learn the multi-day context encoder}
\State Initialize context encoder $C_\phi$ from scratch, including its zero-initialized bias-free map $R_\phi$.
\For{each context update}
  \State $L\gets\Call{Episode}{C_\phi,\mathcal P,H,\ell_{336}}$
  \State $L_o\gets\Call{Episode}{C_\phi,\mathcal P_o,H_o,\ell_o}$ \Comment{separate observed-last-day-state episode}
  \State Update $\phi$ with $\nabla_\phi(L+\lambda_o L_o)$ by backpropagation and AdamW.
\EndFor
\State Fix $C_\phi$, set $R\gets R_\phi$, and retain the base representation $z_{\rm base}=C_\phi(X)$.
\algstage{C}{Learn clock-aware memory}
\State Load fixed $C_\phi,R$; initialize new clock-aware memory parameters $\psi$, zero-initializing only $W$.
\State Set $z_\psi(X)=z_{\rm base}(X)+R(\alpha W u_\psi(X))$.
\For{each clock-aware memory update}
  \State $L\gets\Call{Episode}{z_\psi,\mathcal P,H,\ell_{336}}$.
  \State Update only $\psi$ with $\nabla_\psi L$ by backpropagation and AdamW; keep $C_\phi,R$ fixed.
\EndFor
\algstage{D}{Assemble and export the reported representation}
\State Retain the fixed context and memory learned in Stages B--C on the 8,421-window base pool.
\State Substitute only the daily, dynamics, and physical-scale components adapted on separately selected 9,956- and 13,025-window unlabeled corpora.
\State Re-extract one shared representation and discard all episodic evidence regressors.
\State For each outer fold, select one candidate for all tasks by three-fold inner validation on outer-training participants.
\State Refit each task's StandardScaler+LogReg on outer-training rows; score only its held-out participants.
\alghelper{$\Call{Episode}{f,\mathcal P',H',\ell}$: fit once, then transfer without refitting}
\State Partition eligible recorded identities into disjoint halves; sample 64 endpoints from each half.
\State Encode $Z_F,Z_T\gets f(X_F),f(X_T)$ and derive fixed evidence $E_F,E_T\gets H'(X_F),H'(X_T)$.
\State Fit episodic $\eta_F\gets\operatorname{Ridge}(Z_F,E_F)$ by the differentiable closed-form solve (\ref{eq:ridge}).
\State \Return $\ell(\operatorname{Read}_{\eta_F}(Z_T),E_T)$ without refitting; retain gradients through the fit.
\end{algorithmic}
\end{algorithm}

\paragraph{A: learn base daily representations.}
The daily encoder is trained with masked observable reconstruction, signed profile shape, severity, nested target contrast, and cadence consistency; only observable channels receive the consistency loss. A separately trained auxiliary dynamics encoder uses physical denoising, time-even state functions, time-odd dynamics functions, and variance/covariance regularization. Only its dynamics branch is retained. The base daily encoder and auxiliary dynamics encoder each use 6,960 updates. The daily encoder uses its terminal checkpoint; the auxiliary encoder uses the minimum label-free validation loss at the fixed reporting intervals. Their parameters are then fixed before either multi-day episode loop. Appendix~\ref{sec:daily-implementation} defines the final daily representation.

\paragraph{B: learn the multi-day context encoder.}
The context encoder is the DistributionBarycenter model, initialized from scratch. Histories are sorted by elapsed day and contain at most seven daily representations. Their support-weighted mean is the identity path. For observation-supported 15-minute cells, a shared $15\!\to\!16\!\to\!8$ MLP forms point states. Its inputs are glucose level, positive and negative same-phase residuals, sine and cosine of physical time, nine base-relative 15-minute trajectory values over the preceding 120 minutes, and a complete-trajectory support flag; glucose coordinates are divided by 100. Query/key attention ($12\!\to\!8$ each) retains level, the clock pair, and the nine relative trajectory values, and links each cell only to earlier-day cells. Eligibility and integration mass remain separate masks and weights. A $16\!\to\!8$ message map adds the matched difference. Each day and the complete history are summarized by the eight weighted means and 28 upper-triangular second-order occupation terms. A 16-state GRU receives the 36 daily terms, day mass, and log elapsed gap after exponential state decay. Five weighted per-coordinate quantiles of the point states add 40 distributional terms. Linear maps combine the recurrent and distributional summaries into 36 context coordinates, and $R:\mathbb R^{36}\!\to\!\mathbb R^{146}$ adds their update to the support-weighted daily mean. All masks are applied before attention and pooling; histories never cross recorded identity, cohort, split, cadence, or source.

Both $R$ and the remaining context-encoder parameters update during stage B. The main objective is the 336-coordinate episodic evidence loss of Equation~\ref{eq:outer}. A separate independently sampled episode adds recovery of the already-observed last-day state16 with coefficient $1.75$ in the released model. For this auxiliary pool, retained histories have at least two eligible supported days within the same seven-day observed-span constraint; the target is the state of the final observed input day, not a future day. Its loss averages four groups of state coordinates, with each group weighted equally. Both episodes use the same differentiable, fit-only standardized Ridge regression; the main and auxiliary episode samplers each enforce recorded-identity separation between their own fit and transfer roles.

The released context encoder uses seed 43, 6,960 AdamW updates, learning rate $4\times10^{-4}$, weight decay $10^{-4}$, and gradient clipping at one. Stage C loads the learned context encoder and its normalization parameters, then fixes all inherited parameters, including $R$. The base representation is already history-aware; its comparison with the complete TER system is not a daily-only-versus-history comparison.

\paragraph{Training-seed dispersion.}
Eight seeded representation-training chains (43--50) have seed-specific daily and signed-dynamics branches, recomputed physical-scale adapters, and independently continued context and memory optimization from the same distributed unlabeled initialization. All terminal checkpoints are evaluated on the common outer folds; none is selected by these scores. The sample standard deviations of the three overall metrics are 0.303/0.326/0.239 percentage points. This separate dispersion study uses fixed-checkpoint evaluation; Table~\ref{tab:main} reports training-internal candidate selection followed by outer testing.

\paragraph{C and D: clock-aware memory, final assembly, and linear probing.}
Stage C uses the main 336-coordinate objective only, not the observed-state auxiliary. It initializes the clock-aware memory lift/read/mix parameters and zero-initializes only its writeback $W$; the inherited map $R$ is neither reinitialized nor optimized. Its 10,440-update schedule is described in Appendix~\ref{sec:memory-implementation}. For the final model, the context and memory learned on the 8,421-window base pool remain fixed. Only the local daily branch is adapted for 38,527 label-free updates on 9,956 windows, and the dynamics branch is continued for 15,944 updates on 13,025 windows; the physical-scale component is recomputed from the adapted daily branch. These stage-specific allocations are summarized in Appendix~\ref{sec:data-inventory}. No phenotype label, downstream score, episodic regressor, or episode-fitted scaler enters this assembly. The resulting encoder is then held fixed and the existing downstream StandardScaler/logistic-regression protocol is applied within training folds. This final probe is distinct from both training-only Ridge regressors. The main pretraining fitting and fixed audit partitions retain the boundaries of Appendix~\ref{sec:identity-scope}.

\section{Time interfaces in CGM encoders}
\label{sec:time-interface-comparison}
The distinction in Table~\ref{tab:time-interface} concerns the objects explicitly passed into aggregation, not whether a contextual encoder can learn clock interactions.

\begin{table}[H]\centering
\caption{Time interfaces in CGM. TOD means recorded time of day. The distinction concerns explicit aggregation objects, not the ability of other encoders to learn time interactions.}
\label{tab:time-interface}
\small\setlength{\tabcolsep}{3pt}\renewcommand{\arraystretch}{1.08}
\begin{tabular}{@{}>{\raggedright\arraybackslash}p{.19\linewidth}>{\raggedright\arraybackslash}p{.29\linewidth}>{\raggedright\arraybackslash}p{.47\linewidth}@{}}
\toprule
Model & How time enters & Evidence passed to aggregation \\\midrule
GlucoFM & Circular TOD and patch position & Contextual patch means; post-hoc daily mean or mean--max.\\
CGMformer & Sinusoidal index in midnight-aligned days & Contextual token means; daily representations averaged.\\
CGM-JEPA / X-CGM-JEPA & Position; local calendar marks$^*$ & Single-window contextual patches; paper also studies aligned profiles averaged before encoding.\\
Ours & Raw level/innovation $\times$ physical phase, then elapsed age & Whole-day harmonic slots; learned cross-day mixing added to the base representation.\\
\bottomrule\end{tabular}
\par\smallskip\begin{minipage}{\linewidth}\footnotesize
Sources: \citet{li2026glucofm,lu2025cgmformer,muhammad2026cgmjepa} and audited extraction code. $^*$JEPA defaults disable time features. Calendar marks are passed locally, but their use during released-checkpoint training is unverified. X-JEPA adds glucodensity during training; downstream uses the temporal encoder.
\end{minipage}
\end{table}

\clearpage
\section{Performance across one to seven days of history}
\label{sec:history-length-curve}

Table~\ref{tab:history-length-curve} uses a separate fixed-checkpoint history-length diagnostic with its original task/view protocol.
It averages the same ten cohort--task/view units that support seven days;
Hall's three units support at most four days and remain in the taskwise
endpoint table. The fixed TER checkpoint is unchanged: at each length $K$,
only the most recent $K$ days are retained before extracting its representation.

\begin{table}[ht]\centering\small
\caption{\textbf{Native-history length in the ten seven-day-eligible units.}
Mean task/view score (\%). Higher is better.}
\label{tab:history-length-curve}
\begin{tabular}{rrrr}
\toprule
History $K$ (days) & PR-AUC & ROC-AUC & Macro-F1\\\midrule
1 & 78.04 & 71.11 & 58.79\\
2 & 81.47 & 74.79 & 62.67\\
3 & 82.66 & 76.16 & 64.45\\
4 & \textbf{83.28} & \textbf{76.95} & 65.24\\
5 & 83.04 & 76.82 & 65.29\\
6 & 82.82 & 76.52 & 65.50\\
7 & 82.47 & 75.96 & \textbf{65.90}\\
\bottomrule
\end{tabular}
\end{table}

The largest gains occur when extending one day to the first several days:
PR-AUC and ROC-AUC peak at four days, while Macro-F1 is highest at seven.
This is a practical history-length comparison; as in the native multiday
protocol, longer $K$ changes the eligible start positions. The fixed-endpoint
mean--max and BiLSTM controls in Table~\ref{tab:fixed-history-pooling} separately test how
the same available histories are organized.

\section{Recurring patterns and information selection}
\label{sec:why-details}

This study separates a measured property of real recordings, a controlled input intervention, and a real-label readout test. The generated-signal learners are small matched encoders, not retrained copies of the released three-stage model. Their purpose is to test which information a recovery objective makes accessible.

\begin{figure}[htbp]\centering
\includegraphics[width=\linewidth]{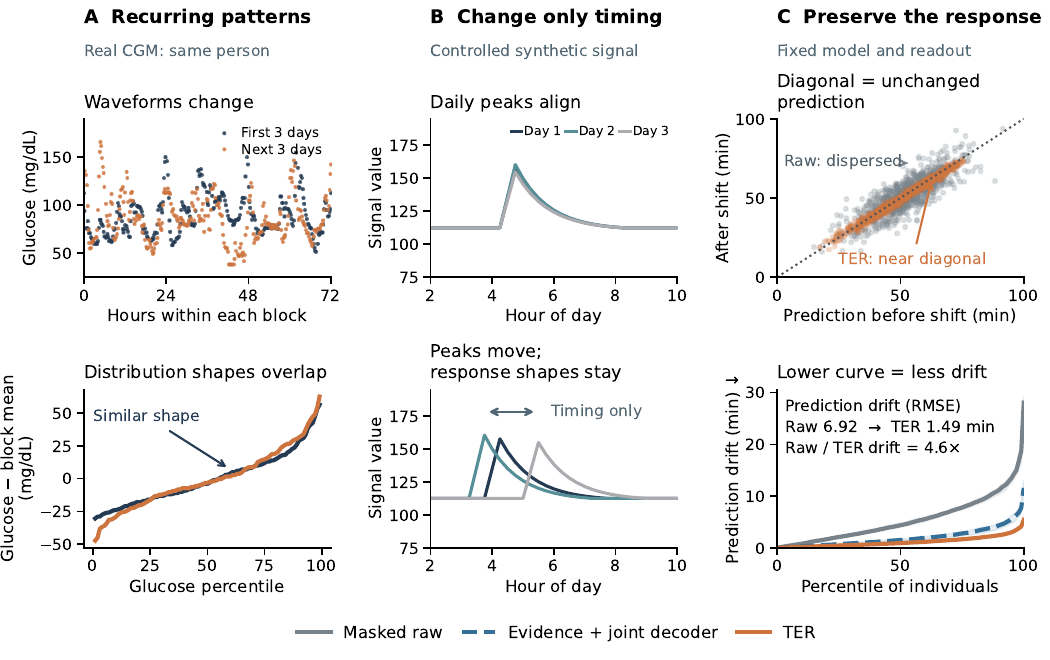}
\caption{\textbf{Variable recordings, repeatable patterns.} (A) Two real three-day blocks from one preselected person. (B) Controlled seven-day signals change event timing while preserving response shapes and values. (C) Fixed encoders and readouts predict an unchanged response timescale before/after shifts of up to 60 minutes. Top: all 1,024 test individuals, one prespecified seed; diagonal means unchanged prediction. Bottom: drift quantiles, mean/range over three seeds.}
\label{fig:why}
\end{figure}

\subsection{Real, nonoverlapping date blocks}

We select the earliest eligible six consecutive days per biological participant, keeping acquisition source and cadence consistent: 44 CGMacros, 49 Shanghai, 33 Stanford, and 11 Hall participants (137 total). Blocks A and B comprise the first and next three days. Observation masks determine native-cadence coverage; missing dates are not interpolated or joined. Descriptive repeatability uses observed-only 15-minute bins and the intersection of valid bins in each pair. Each block is centered before calculating either time-aligned squared distance or sorted-value squared distance. Each distance is divided by its own mean distance to other participants matched on cohort, source, cadence, and coverage (within .1); at least two controls are required. One CGMacros participant lacks such controls, leaving 136 for this description. The smaller distribution ratio is not inferred from the mechanically smaller absolute distance after sorting: each measure has a separate between-person denominator.

\begin{table}[ht]\centering\small
\caption{Median same-person/between-person distance ratios. Lower indicates more repeatable participant structure relative to matched controls.}
\begin{tabular}{lrrr}\toprule
Cohort & People & Aligned traces & Distribution shape\\\midrule
CGMacros & 43 & .667 & .099\\
Shanghai & 49 & .746 & .337\\
Stanford & 33 & .902 & .466\\
Hall & 11 & .496 & .174\\\bottomrule
\end{tabular}
\end{table}

The illustrated CGMacros participant is nearest the joint median rank of the two ratios, selected before inspecting model performance. Both blocks are shown without filling missing samples. A separate normalized-change-profile diagnostic measures RMS change at 15/30/60/120 minutes divided by $\sqrt{2}$ times within-block SD, excluding changes across gaps. The corresponding ratios are .196/.169/.380/.340. These are observational repeatability results, not causal claims about phenotype labels.

\subsection{Matched recovery objectives on controlled signals}

Each simulated participant has seven 96-point days, three responses per day, a baseline sampled uniformly from 80--120, amplitude 25--60, and a response-shape parameter $\tau$ sampled uniformly from 2--5 time steps. Samples are 15 minutes apart. Event starts are near samples 16/44/72, with a participant-specific offset of up to two samples. For $k=0,\ldots,16$, each compact pulse is
\[
h_\tau(k)=\begin{cases}
k/2,& k\leq2,\\
\bigl[\exp(-(k-2)/\tau)-\exp(-14/\tau)\bigr]/\bigl[1-\exp(-14/\tau)\bigr],&k>2.
\end{cases}
\]
Event amplitudes vary by a factor of .8--1.2. Additional event/day timing variation is 0, up to 30, or up to 60 minutes. The same participant parameters and event amplitudes are reused across conditions. Pulses never overlap or cross day boundaries, so moving them preserves the exact multiset of observed values, peaks, response shapes, and total load. The response target $15\tau$ is a generated shape timescale, not a clinical recovery half-life. The second target is the three event start times on the final day.

Independent populations contain 4,096 pretraining, 256 readout-fit, and 1,024 test participants. The 28 unlabeled evidence targets are 16 quantiles, four RMS changes at lags 1--4, four fractions exceeding the within-record median by 10/20/30/40, and mean/median/SD/95th-minus-5th percentile. Evidence is computed only from the input signal, never from the generated response parameter. These summaries are not all strictly invariant to the timing intervention.

All methods use the same 101,088-parameter encoder: two stride-2 convolutions (32/64 channels, kernel 7), GELU, adaptive pooling to 21 positions, and a 64-output linear layer with LayerNorm. Each batch has 128 distinct people; 25\% of four-sample patches are masked. Initialization, batch and mask sequences, and 2,000 updates match across methods within each seed (43--45). AdamW uses learning rate .001, weight decay .0001, and gradient clipping at 5. Last-step checkpoints are used, with no condition, seed, or checkpoint selection.

Masked raw recovery trains a linear decoder on masked samples only. Joint evidence reconstruction trains a linear decoder jointly with the encoder on standardized evidence. TER splits the batch into 64 fit and 64 transfer participants and differentiates through a centered Ridge solve with penalty 1. Its temporary reader is discarded after each update. Encoder capacity and update budget match; the training-only output heads differ with their objectives. After training, the same standardized Ridge readout protocol predicts each evaluation target. Table~\ref{tab:why-timing} includes all conditions.

\begin{table}[ht]\centering\small
\caption{Controlled task-dependent information selection. Normalized MSE, mean over three training seeds. Each encoder and readout is trained at its stated timing condition. Lower is better.}
\label{tab:why-timing}
\begin{tabular}{lrrrrrr}\toprule
 & \multicolumn{3}{c}{Response timescale} & \multicolumn{3}{c}{Exact event timing}\\
Maximum timing variation & 0 & 30 & 60 & 0 & 30 & 60\\\midrule
Masked raw recovery & .2813 & .5118 & .6348 & .4642 & .7436 & .7662\\
Joint evidence reconstruction & .1622 & .1482 & .1512 & .7849 & .9108 & .9306\\
TER & .1116 & .0963 & .1037 & .8143 & .9703 & .9567\\
Direct evidence & .0238 & .0238 & .0238 & 1.0512 & 1.0392 & 1.0268\\
Direct raw input & .1230 & .3781 & .6981 & .0775 & .1949 & .2482\\\bottomrule
\end{tabular}
\end{table}

The raw-minus-TER response-error gap grows from the zero- to 60-minute condition by .3527/.3281/.4035 across seeds. Joint evidence also retains response information; TER improves on it in all nine seed--condition pairs. Raw recovery predicts exact timing better in all nine pairs. Direct evidence is a stronger reference for the response parameter than any learned representation in this constructed setting: the study establishes information selection, not superiority over sufficient statistics.

\subsection{Fixed-model intervention shown in Figure~\ref{fig:why}}

We reuse the nine encoders trained without additional timing variation. Each downstream reader is fitted once on 256 zero-variation participants; the encoder, scaler, intercept, and coefficients remain unchanged while the same 1,024 test individuals receive 0/30/60-minute timing variation. Thus neither representation adaptation nor readout refitting can explain the paired prediction changes. Table~\ref{tab:why-fixed} reports the largest intervention. Normalization uses the zero-variation fitting population throughout; these errors should not be mixed with condition-specific normalization in Table~\ref{tab:why-timing}.

\begin{table}[ht]\centering\small
\caption{Fixed encoder and readout, before/after timing changes. Prediction drift is RMSE between predictions before and after, in simulated timescale minutes (mean $\pm$ seed SD).}
\label{tab:why-fixed}
\begin{tabular}{lrrrr}\toprule
 & \multicolumn{2}{c}{Response MSE} & Response drift & Timing MSE\\
Method & Before & After & After vs. before & After\\\midrule
Masked raw & .2813 & .5043 & $6.915\pm.448$ & 3.3366\\
Joint evidence & .1622 & .1681 & $2.642\pm.315$ & 3.5530\\
TER & .1116 & .1127 & $1.490\pm.108$ & 3.6416\\\bottomrule
\end{tabular}
\end{table}

TER's response-prediction correlations with ground truth after the intervention are .939/.939/.936. Low drift therefore does not reflect a constant output. The main scatter uses the prespecified seed43 and every test individual; the quantile plot covers 0--100\% of people with mean and min--max range across seeds. The annotated 4.6 ratio is the ratio of seed-mean drift RMSEs. Exact timing actually changes, so low timing-prediction drift would not itself be desirable. This task-dependent distinction is compatible with retaining clinically informative clock associations in the CGM encoder.

\subsection{Readout on independently observed real days}

The two existing four-arm sets use the original daily input and a later daily-input retraining, respectively. Within each set, raw-hourly/structured targets are crossed with joint decoding/episodic Ridge regression. Each A representation reads only A days; each B representation reads only B days. Labels are never read during representation extraction. For the 14 phenotypes, ten repetitions of five participant-grouped stratified folds fit StandardScaler plus L2 logistic regression ($C=1$) on training participants' A representations. The unchanged classifier predicts held-out participants from A and B. No B adaptation or hyperparameter search is performed.

Probability drift is $\frac12\sum_c|p_{B,c}-p_{A,c}|$; label error is $\frac12\sum_c(p_c-\mathbf1[y=c])^2$. In binary tasks these are absolute positive-probability change and Brier score. A training-prevalence predictor has zero drift but nonzero label error, so stability is assessed together with accuracy. Table~\ref{tab:why-date} averages tasks equally.

\begin{table}[ht]\centering\small
\caption{Natural date replacement: all four objectives in both existing training sets. Lower is better. These subset diagnostics are distinct from main-table benchmark scores.}
\label{tab:why-date}
\begin{tabular}{llrrr}\toprule
Daily input & Training target/reader & Drift & A label error & B label error\\\midrule
Original & Raw/joint & .233342 & .287777 & .293440\\
 & Raw/episodic & .105824 & .212370 & .210677\\
 & Evidence/joint & .225600 & .296989 & .293483\\
 & Evidence/episodic & .107704 & .208778 & .205782\\\midrule
Later & Raw/joint & .206883 & .270412 & .265086\\
 & Raw/episodic & .105123 & .202166 & .200276\\
 & Evidence/joint & .210583 & .269370 & .263948\\
 & Evidence/episodic & .100931 & .210844 & .204198\\\bottomrule
\end{tabular}
\end{table}

With structured targets fixed, episodic fitting versus joint decoding improves drift on 13/14 and 12/14 tasks, B-label error on 14/14 and 12/14, and both on 13/14 and 10/14. Participant-paired descriptive 95\% bootstrap intervals exclude zero for drift in 10/14 and 9/14 tasks, and for B-label error in 5/14 and 6/14; these intervals are not multiplicity-adjusted. Hall has only 11 eligible people and 3--4 members in some minority classes. Training folds retain both classes, and Brier/drift remain defined for one-class test folds.

Holding episodic fitting fixed and changing raw to evidence improves B-label error on 9/14 and 7/14 tasks, but improves error and drift together on only 3/14 in either set. Raw targets also benefit from episodic fitting. Separately, A-to-B normalized-change-profile readouts favor structured/episodic over raw/episodic in mean in all eight cohort--set comparisons (1.82--61.95\%); only the two Shanghai paired intervals exclude zero. Prespecified natural waveform/profile variability strata do not show a consistent monotonic benefit. Natural date changes can contain genuine physiological changes, unlike the controlled timing intervention.

The original full-history target--reader factorial favors structured/episodic over raw/episodic by 2.883/3.234/2.230 PR-AUC/ROC-AUC/Macro-F1 points. Repeating the comparison with the later daily input gives differences of $-.851/-.861/-1.495$ points; a fine-raw target comparison is also unfavorable in aggregate. These are distinct training trajectories, not repeated estimates of the released checkpoint. They motivate the conditional information-selection interpretation rather than attribution of all phenotype gains to timing invariance. Readout participant separation and A/B date separation do not imply that the public participants were absent from representation pretraining.

\section{Supplementary benchmark and evidence summaries}
\begin{figure}[H]\centering
\includegraphics[width=.94\linewidth]{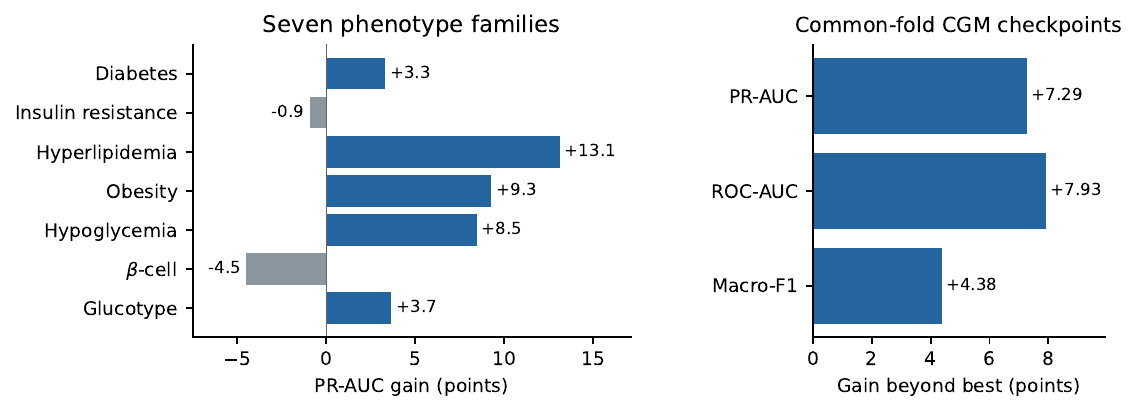}
\caption{\textbf{Breadth and magnitude on the adopted evaluation.} Left: TER's PR-AUC difference from the strongest CGM model in each phenotype family. Right: gains beyond the strongest of the three locally evaluated CGM checkpoints on common outer folds.}
\label{fig:phenotype-family-breadth}\label{fig:checkpoint-gap}
\end{figure}
\begin{center}
\begin{minipage}{\linewidth}
\centering
\textbf{Evidence used for reconstruction (336 values)}
\par\smallskip
\begin{minipage}[c]{.29\linewidth}
\centering
\begin{tikzpicture}
  \draw[stageB,line width=8pt] (90:.58) arc[start angle=90,end angle=158.57,radius=.58];
  \draw[stageC,line width=8pt] (158.57:.58) arc[start angle=158.57,end angle=432.86,radius=.58];
  \draw[stageA,line width=8pt] (432.86:.58) arc[start angle=432.86,end angle=450,radius=.58];
  \node[align=center] at (0,0) {\textbf{336}\\[-2pt]\scriptsize evidence\\[-2pt]\scriptsize values};
\end{tikzpicture}
\end{minipage}\hfill
\begin{minipage}[c]{.67\linewidth}
\footnotesize
\textcolor{stageB}{\rule{1.1ex}{1.1ex}}\enspace\textbf{Current-day (64).} State, rise/recovery excursions, extremes, and rhythm/observation.\\[2pt]
\textcolor{stageC}{\rule{1.1ex}{1.1ex}}\enspace\textbf{Longitudinal (256).} Mean, spread, directed extreme, and last--first change of each current-day evidence value.\\[2pt]
\textcolor{stageA}{\rule{1.1ex}{1.1ex}}\enspace\textbf{Low/high events (16).} Occurrence, persistence, onset/offset, and burden across days.\\[3pt]
\emph{MAE recovers masked measurements; TER reconstructs this evidence using a regressor fitted on other recordings.}
\end{minipage}
\end{minipage}
\end{center}

\clearpage
\section{Transfer from private-only pretraining}
\label{sec:private-only-transfer}

\paragraph{Question and setup.}
Can TER transfer to public cohorts without public CGM in pretraining?
We remove public rows from every stage-specific pool and retrain all
components using only the de-identified, user-authorized private corpus
(Appendix~\ref{sec:data-inventory}). Normalization and evidence statistics
are fitted on private training partitions; no public pretrained weight or
phenotype label is used. Architecture, objectives, budgets, and seed 43
are fixed before evaluation, without downstream checkpoint selection.

The base pool contains 6,138 daily windows (5,473 fitting/665 validation).
Final daily, dynamics-initialization, and final dynamics pools contain
7,673, 7,673, and 10,742 windows; these subsets overlap. Base daily/dynamics
training uses 6,960 updates each, context/memory uses 6,960/10,440,
and final daily/dynamics uses 38,527 and 9,965+15,944.
Daily/context/memory use terminal checkpoints; dynamics uses fixed
label-free validation. Fitting and validation identities remain disjoint.

\paragraph{Evaluation.}
We freeze the representation and apply Table~\ref{tab:main}'s fourteen-task,
subject-disjoint $10\times5$ protocol, with identical fold fingerprints
and labels. StandardScaler and logistic regression fit only training folds.
Tables~\ref{tab:private-only-summary}--\ref{tab:private-only-tasks} report
aggregate and task-level results.

\begin{table}[H]\centering\small
\caption{Pretraining-source comparison on common outer folds (\%). Single-source variants are fixed trained checkpoints; the main row uses the training-internal selector.}
\label{tab:private-only-summary}
\setlength{\tabcolsep}{4pt}
\begin{tabular*}{\linewidth}{@{\extracolsep{\fill}}lrrr@{}}
\toprule
Representation & PR-AUC & ROC-AUC & Macro-F1\\\midrule
Public only & 58.88 & 67.63 & 57.25\\
Private only & 60.73 & 68.74 & 58.32\\
Public + private (main table) & 63.75 & 71.13 & 60.56\\
\bottomrule\end{tabular*}
\end{table}

\paragraph{Results.}
Private-only TER improves public-only TER by 1.85/1.11/1.07 points and reaches 8/14 best-metric tasks. Combined-source TER improves private-only TER by 3.02/2.38/2.24 points. These results test transfer from separate unlabeled data sources under the same downstream evaluation.

The private base pool supplies 77 eligible multi-day fitting endpoints
from 54 identities and ten validation endpoints from ten identities
(median history: three days). Observation and identity constraints are
unchanged; only single-source support checks are adapted. This tests
the available private-source histories, not equal-size cohort substitution.

\begin{table}[H]\centering\footnotesize
\caption{Private-only TER scores and differences from the combined-source main result (percentage points).}
\label{tab:private-only-tasks}
\setlength{\tabcolsep}{4pt}
\begin{tabular*}{\linewidth}{@{\extracolsep{\fill}}lrrrrrr@{}}
\toprule
Task & PR & ROC & F1 & $\Delta$ PR & $\Delta$ ROC & $\Delta$ F1\\\midrule
CGMacros Diabetes & 69.76 & 80.94 & 58.77 & -1.60 & -0.90 & -1.75\\
CGMacros Insulin resistance & 93.04 & 87.36 & 77.20 & 0.59 & 0.51 & -1.19\\
CGMacros Hyperlipidemia & 31.58 & 50.90 & 46.60 & -12.69 & -6.50 & -7.41\\
CGMacros Obesity & 72.19 & 70.84 & 60.41 & -3.53 & -3.47 & -3.13\\
Shanghai Insulin resistance & 53.39 & 34.08 & 33.43 & -3.71 & -4.76 & -3.34\\
Shanghai Hyperlipidemia & 41.21 & 54.88 & 49.59 & -7.24 & -9.79 & -4.44\\
Shanghai Hypoglycemia & 26.21 & 67.91 & 50.19 & -3.34 & 2.36 & 0.23\\
Stanford Diabetes & 83.96 & 82.02 & 68.28 & 2.61 & 0.94 & 0.52\\
Stanford Beta-cell dysfunction & 64.62 & 65.63 & 54.12 & 0.11 & -0.66 & -2.18\\
Stanford Insulin resistance & 66.87 & 70.83 & 56.19 & -1.96 & -2.40 & -4.70\\
Hall Diabetes & 64.65 & 75.23 & 62.71 & -1.94 & -0.33 & -1.76\\
Hall Insulin resistance & 67.25 & 73.72 & 65.63 & 1.96 & 0.46 & 0.94\\
Hall Hyperlipidemia & 23.94 & 53.95 & 47.98 & -11.16 & -8.95 & -2.87\\
Hall Glucotype & 91.59 & 94.10 & 85.37 & -0.37 & 0.16 & -0.29\\
\bottomrule\end{tabular*}
\end{table}

\clearpage
\section{Direct features and the estimation of new readouts}
\label{sec:readout-mechanism}

The main-table-aligned phenotype comparisons below use the common outer partitions and retain training-internal selection for multi-candidate arms. The whitening and observation-perturbation diagnostics retain their original fixed-model partitions, as specified in their subsections. The label-free common-reader experiments use fixed checkpoint representatives and 24 person-grouped audit splits to measure recovery error and fit-group variance. Paired-seed interventions and coordinate controls specify their fixed model pairs.

\subsection{Does retaining daily information repair direct evidence features?}

We concatenate the retained daily/dynamics representation with the same aligned
336 evidence coordinates, giving 482 features. There is no learned feature
selection or additional pretraining. Window order and recorded identities are
checked against the released representation; phenotype labels are used only
for the fold-local downstream classifier. All 14 tasks use the main
evaluation's ten-by-five outer folds, StandardScaler, and logistic regression, with matching
fold fingerprints. Table~\ref{tab:direct-evidence-complement} gives the result.

\begin{table}[H]\centering\small
\caption{Providing evidence is not equivalent to learning with it. Scores on common outer folds (\%).}
\label{tab:direct-evidence-complement}
\setlength{\tabcolsep}{4pt}
\begin{tabular*}{\linewidth}{@{\extracolsep{\fill}}lrrr@{}}
\toprule
Representation & PR-AUC & ROC-AUC & Macro-F1\\\midrule
Daily only & 54.89 & 64.48 & 56.68\\
Direct evidence & 56.49 & 64.55 & 56.97\\
Daily + evidence & 56.49 & 65.41 & 57.02\\
TER (main table) & 63.75 & 71.13 & 60.56\\
\bottomrule\end{tabular*}
\end{table}

Retaining daily information leaves a 7.26/5.72/3.54-point gap to TER. Thus
removing the daily representation is not by itself the explanation for the
direct-evidence result. Evidence targets guide learning without fixing the
exported representation to their original coordinates. This comparison does
not isolate summary incompleteness, nonlinear organization, and estimation
complexity from one another, nor does it measure sensor noise.

\subsection{Common-reader fitting and transfer}

We use all 2,708 windows from 45/58/37/56 biological participants in
CGMacros/Shanghai/Stanford/Hall (196 total). The four representations share
identical rows and a fixed, pretraining-normalized evidence bank. For each of
24 paired random splits within each cohort, 67\% of people fit a StandardScaler
and Ridge regressor ($\alpha=1$); the remaining 33\% evaluate it. Each person's
windows have total fitting weight one. Errors are averaged by person and then
equally across cohorts. The longitudinal/event group contains 256 across-day
coordinates and 16 event coordinates; current-day evidence contains 64.

\begin{table}[ht]\centering\small
\caption{\textbf{Fit accuracy versus reliability of a new readout.} Longitudinal/event
MSE and prediction variance; lower is better. The two rightmost columns use
fixed evaluation people and repeated fit-group draws, separately from the
67/33 splits in the middle columns.}
\label{tab:canonical-readout-stability}
\begin{tabular}{llrrrr}\toprule
Target & Objective & Fit MSE & Held-out MSE & Fixed-group MSE & Variance\\\midrule
Evidence & Joint & .22828 & 1.02079 & .86516 & .15048\\
Evidence & Fit--transfer & .51716 & .93548 & .80251 & .04290\\
Raw & Joint & .26896 & 1.07450 & .94115 & .15848\\
Raw & Fit--transfer & .60486 & 1.02598 & .90703 & .04660\\\bottomrule
\end{tabular}
\end{table}

The stability test fixes 15/20/13/19 evaluation people by cohort and draws
24 fitting groups of 20/25/16/24 people from the remaining participants.
Every objective variant uses the same draws. Variance measures changes in
predictions on those fixed evaluation people, not changes in coefficients
across differently parameterized embeddings or training-seed variance.
For structured targets, TER reduces this variance by 71.49\% and the separate
67/33-split held-out error by 8.36\%. Current-day reductions are 78.57\% and
1.18\%, respectively.

\begin{table}[ht]\centering\small
\caption{\textbf{Longitudinal/event readout across all four cohorts.}
Held-out MSE uses 67/33 splits; variance reduction uses fixed evaluation people.}
\begin{tabular}{lrrr}\toprule
Cohort & Joint MSE & TER MSE & Variance reduction (\%)\\\midrule
CGMacros & 1.85643 & 1.85294 & 34.33\\
Shanghai & .90762 & .86571 & 69.56\\
Stanford & .58704 & .55365 & 73.31\\
Hall & .73206 & .46962 & 86.56\\\bottomrule
\end{tabular}
\end{table}

For fixed evaluation targets $e$, let $\widehat e_j$ be predictions from fit
draw $j$ and $\overline e$ their average. The finite-draw identity
\[
\mathbb E_j\|\widehat e_j-e\|^2
=\|\overline e-e\|^2+\mathbb E_j\|\widehat e_j-\overline e\|^2
\]
holds coordinatewise and under our averaging. Joint reconstruction gives
$.71468+.15048=.86516$; TER gives $.75961+.04290=.80251$.
Here reduced fit-group sensitivity offsets the larger error of the average
prediction. This is a decomposition of the diagnostic, not a new downstream
ensemble or a population bias--variance theorem. Fixed-group total error is
not lower in every cohort: CGMacros has lower variance but higher total error.

\subsection{What is readable from generic history representations?}

The same 24 biological-person splits and Ridge protocol compare the daily,
mean--max, and learned BiLSTM representations with TER. The generic poolers
use the existing matched-history controls; the historical TER row retains
their original assembly comparison, and the released row links the audit to
Table~\ref{tab:main}. All rows use the identical evidence bank. Longitudinal
targets separate across-day mean, spread, signed tail, and last--first change
(64 coordinates each); event targets contain 16 coordinates.

\begin{table}[ht]\centering\small
\caption{\textbf{Evidence recovery on other people.} Held-out MSE, lower is better.
All evidence groups are included.}
\label{tab:history-property-readout}
\begin{tabular}{lrrrrrr}\toprule
Representation & Current & Mean & Spread & Tail & Change & Event\\\midrule
Daily & .635 & 1.085 & 1.182 & 1.849 & 1.193 & 1.083\\
Mean--max & .941 & .970 & .888 & 1.588 & 1.458 & 1.085\\
BiLSTM + daily & .778 & 1.058 & .854 & 1.692 & .922 & 1.025\\
TER (historical assembly) & .625 & .800 & .541 & 1.632 & .748 & .622\\
TER (released) & .632 & .807 & .585 & 1.649 & .771 & .657\\\bottomrule
\end{tabular}
\end{table}

\begin{table}[ht]\centering\small
\caption{\textbf{The generic representations permit accurate fitting.}
Fit-person MSE under the same protocol.}
\begin{tabular}{lrrrrrr}\toprule
Representation & Current & Mean & Spread & Tail & Change & Event\\\midrule
Mean--max & .224 & .086 & .140 & .095 & .355 & .179\\
BiLSTM + daily & .109 & .081 & .099 & .081 & .144 & .110\\
TER (released) & .429 & .405 & .459 & .548 & .664 & .486\\\bottomrule
\end{tabular}
\end{table}

Mean, spread, change, and event held-out errors favor released TER over BiLSTM
in all four cohorts, with overall reductions of 23.77/31.43/16.40/35.92\%.
Tail differences are smaller and mixed; mean--max has lower overall tail error.
The generic representations can fit these properties well, so an inability
to model temporal information is not the appropriate explanation. The audit
instead identifies a gap in reliable estimation of a new readout. Feature
dimensions differ (292 for mean--max, 402 for BiLSTM, 146 for TER), and a common
Ridge penalty need not give equal effective regularization. The following
intervention tests how much this accounts for the observed difference.

\subsection{Can readout regularization repair the difference?}
\label{sec:nested-readout}

We keep all six representations fixed and repeat the same 24 outer person
splits. Within each outer fitting group, three person-disjoint folds choose
one Ridge penalty from $\{.01,.1,1,10,100,1000,10000\}$ for all 336 targets.
Each inner fold fits its own subject-balanced scaler and regressor. Selection
minimizes per-person validation error, equally weighting current-day and
longitudinal/event targets. The selected regressor is refitted on all outer
fitting people before evaluation on the untouched outer group. Neither
phenotype labels nor outer evidence targets enter selection.

\begin{table}[ht]\centering\small
\caption{\textbf{Independent regularization improves controls but leaves a readout gap.}
Held-out evidence MSE. Nested selection uses only outer-fitting participants.}
\label{tab:nested-readout}
\begin{tabular}{lrrrr}\toprule
 & \multicolumn{2}{c}{Current day} & \multicolumn{2}{c}{Longitudinal/events}\\
Representation & $\alpha=1$ & Nested & $\alpha=1$ & Nested\\\midrule
Evidence + fit--transfer (TER) & .63222 & .60590 & .93548 & .89428\\
Evidence + joint & .63974 & .62689 & 1.02079 & .96611\\
Raw + fit--transfer & .72066 & .64112 & 1.02598 & .95168\\
Raw + joint & .71667 & .69965 & 1.07450 & 1.00807\\
Mean--max & .94083 & .78636 & 1.21762 & 1.04371\\
BiLSTM + daily & .77821 & .64503 & 1.12522 & .95783\\\bottomrule
\end{tabular}
\end{table}

Regularization reduces BiLSTM's longitudinal/event error by 14.88\%, compared
with 4.40\% for TER. TER still has 7.44/6.63/14.32\% lower error than joint
evidence reconstruction, BiLSTM, and mean--max; each comparison favors TER
in all four cohorts. The corresponding cohort errors for TER/joint/BiLSTM
are 1.76161/1.87169/1.86620, .83800/.87508/.92603,
.50790/.58234/.54275, and .46962/.53535/.49634.

Across the 96 cohort/split fits, TER selects penalties .01/.1/1 in 13/59/24
cases; joint evidence selects 1/10/100 in 36/56/4 cases. BiLSTM selects
10/100 in 69/27 cases and mean--max in 57/39. Raw fit--transfer selects
.01/.1/1 in 61/25/10 cases, while raw joint selects 1/10/100 in 16/73/7.
The prespecified grid is unchanged despite lower-bound selections. These
overlapping splits give descriptive averages, not 96 independent observations.
The fixed-$\alpha$ rerun reproduces the preceding audit to numerical precision.

The intervention shows that a fixed penalty accounts for part of the original
gap, while a representation difference remains under nested selection.
Section~\ref{sec:fit-transfer-intervention} separately tests refitting the training
reader versus scoring it on different people. The main phenotype classifier
and all released-model results remain unchanged.

\subsection{Changing the objective of the same recurrent encoder}
\label{sec:bilstm-objective-repair}

To test whether the generic encoder's objective contributes to the gap, we
train two copies of the original one-layer bidirectional LSTM (128 units per
direction). Both receive the same cached daily representations and day ages,
use the same causal histories of up to seven days, and export the history
representation together with the unchanged daily representation. The corpus
contains 8,421 windows, with 2,699 training and 457 validation endpoints meeting
the evidence-support requirement. The 336 evidence targets match those used
by the released TER model. Both copies share encoder initialization, seed 43,
6,960 AdamW updates, learning rate .001, weight decay .0001, gradient clipping
at 1, and the same two identity-disjoint groups of 64 endpoints per update.
We evaluate the last checkpoint, without model selection.

The reconstruction arm uses the original age-conditioned decoder to recover
the observed daily representations in both groups. The fit--transfer arm
instead fits a differentiable Ridge readout on one group and recovers evidence
on the other, using TER's current-day and longitudinal/event weighting.
Neither arm receives the clock module or TER's auxiliary training loss.
The recurrent encoder has 283,648 parameters in both arms; reconstruction
additionally optimizes its decoder (387,218 total parameters), whereas
fit--transfer solves its reader analytically. Thus this intervention changes
the learning objective, including its target and reader, rather than target
choice alone. It uses the original BiLSTM daily inputs, not TER's later daily
checkpoint, and is a within-architecture comparison.

\begin{table}[H]\centering\footnotesize
\caption{BiLSTM evidence fit--transfer minus daily reconstruction on every task (points).}
\label{tab:bilstm-task-delta}
\setlength{\tabcolsep}{4pt}
\begin{tabular*}{\linewidth}{@{\extracolsep{\fill}}lrrr@{}}
\toprule
Task & $\Delta$ PR & $\Delta$ ROC & $\Delta$ F1\\\midrule
CGMacros Diabetes & -4.65 & -4.01 & -5.59\\
CGMacros Insulin resistance & 0.45 & 0.84 & 0.01\\
CGMacros Hyperlipidemia & 1.66 & 1.01 & 0.56\\
CGMacros Obesity & 3.77 & 6.83 & 6.50\\
Shanghai Insulin resistance & -6.01 & -10.75 & -7.43\\
Shanghai Hyperlipidemia & -6.62 & -7.51 & -3.80\\
Shanghai Hypoglycemia & -3.39 & -2.95 & -1.39\\
Stanford Diabetes & -1.88 & -0.44 & 0.42\\
Stanford Beta-cell dysfunction & 3.30 & 2.91 & 0.44\\
Stanford Insulin resistance & 3.36 & 3.07 & 3.34\\
Hall Diabetes & 4.21 & 3.12 & 4.24\\
Hall Insulin resistance & -4.29 & -5.22 & -2.87\\
Hall Hyperlipidemia & 10.04 & 9.77 & 4.01\\
Hall Glucotype & 5.88 & 6.75 & 7.06\\
\bottomrule\end{tabular*}
\end{table}

\subsection{Does a freshly fitted reader still need cross-person scoring?}
\label{sec:fit-transfer-intervention}

We train three paired runs (seeds 43, 44, 45) with the released model's daily
inputs, 8,421-window corpus, targets, architecture, optimizer, and budgets
(6,960 multi-day and 10,440 memory updates). Within each pair, initialization
and sampled groups are identical. Each update solves two regularized readers
$W_S$ and $W_Q$ from the same identity-disjoint groups $S,Q$, including the
same normalization and gradients through each solve. The fit-scored arm
minimizes $\tfrac12[\ell(Z_SW_S,E_S)+\ell(Z_QW_Q,E_Q)]$; the transfer-scored
arm minimizes $\tfrac12[\ell(Z_QW_S,E_Q)+\ell(Z_SW_Q,E_S)]$.
We apply this distinction to the main and auxiliary multi-day objectives and
the memory objective. Both arms retain the original other-person validation
procedure and use the last checkpoint without selecting among seeds or steps.
Both use symmetric scoring, so the comparison is between these new paired
runs, not between one new control and the published one-direction model.

\begin{table}[H]\centering\footnotesize
\caption{Cross-person minus within-group scoring, averaged over all three paired seeds (points).}
\label{tab:fit-transfer-task-delta}
\setlength{\tabcolsep}{4pt}
\begin{tabular*}{\linewidth}{@{\extracolsep{\fill}}lrrr@{}}
\toprule
Task & $\Delta$ PR & $\Delta$ ROC & $\Delta$ F1\\\midrule
CGMacros Diabetes & 4.85 & 2.00 & 2.58\\
CGMacros Insulin resistance & 0.74 & 1.23 & 4.02\\
CGMacros Hyperlipidemia & 7.03 & -0.19 & 1.96\\
CGMacros Obesity & 8.49 & 8.64 & 4.45\\
Shanghai Insulin resistance & -4.86 & -9.49 & -8.52\\
Shanghai Hyperlipidemia & 5.80 & 0.43 & 0.66\\
Shanghai Hypoglycemia & 7.05 & 5.57 & -1.06\\
Stanford Diabetes & 6.27 & 6.71 & 4.13\\
Stanford Beta-cell dysfunction & 6.73 & 9.14 & 4.68\\
Stanford Insulin resistance & 9.64 & 9.56 & 5.15\\
Hall Diabetes & 15.05 & 13.21 & 10.98\\
Hall Insulin resistance & 15.49 & 10.82 & 9.44\\
Hall Hyperlipidemia & 11.98 & 7.01 & 4.02\\
Hall Glucotype & 13.42 & 12.41 & 13.44\\
\bottomrule\end{tabular*}
\end{table}

\subsection{Can an information-preserving coordinate change repair direct features?}
\label{sec:evidence-geometry}

For direct evidence, daily plus evidence, and released TER, we test one
prespecified full-rank shrinkage-whitening transformation. Within each
downstream training fold, we fit the original StandardScaler and estimate
Ledoit--Wolf covariance from standardized training rows. Its symmetric inverse
square root transforms the features; eigenvalues are floored at $10^{-10}$
for numerical invertibility. One scalar preserves the original training mean
squared feature norm. No labels or held-out rows fit this transformation;
no dimensions are removed and no second coordinate-wise scaling is applied.
The L2 logistic classifier, folds, and scoring remain fixed. This diagnostic
changes preprocessing and is not a replacement for the main-table protocol.

This independent fixed-checkpoint preprocessing audit retains its original
task-stratified folds. Whitening reduces PR-AUC by 3.89, 4.80 and 7.63 points
for direct evidence, concatenation and TER, respectively. It is not a
re-evaluation of the main table's nested selection procedure.

Training PR-AUC increases from 99.49 to 99.83\% for evidence, 99.92 to
100.00\% for daily plus evidence, and 85.06 to 98.89\% for TER, while held-out
performance decreases. All transforms pass inversion and norm-preservation
checks. In particular, TER's information is unchanged but its predictive
utility under the fixed classifier falls: an invertible map can change the
effective regularization of a finite-sample linear readout. The result rules
out simple whitening as a repair, not every alternative representation of
the summaries. It neither removes sensor noise nor restores information
absent from the summaries, so those explanations remain distinct.

\subsection{Can a source-learned affine map make the same summaries more useful?}
\label{sec:summary-projection}

We distinguish information availability from its usefulness to a finite-sample
classifier through a fixed input-restriction experiment. Using the released TER
encoder as a label-free teacher, we predict its representation from three
prespecified inputs: direct evidence, daily representations, and their
concatenation. Each arm fits one StandardScaler and Ridge regressor
($\alpha=1$, with intercept) on 7,341 pretraining fitting windows from 4,299
recorded identities. The remaining 1,080 windows from 539 disjoint recorded
identities are used only for reconstruction audit. Source fitting rows determine
both input and teacher normalization; inverse identity-frequency weights sum
to the number of fitting rows. No phenotype labels select or fit the map.
All arms export 146 coordinates and use the unchanged subject-grouped
$10\times5$ StandardScaler--LogisticRegression evaluation.

The pretraining teacher uses the original daily input; public evaluation uses
the final daily input of the released assembly. Before fitting, we recomputed
the teacher and exact evidence on all four public cohorts: both matched the
archived arrays elementwise. The source-fitted maps are unchanged; their downstream scores use the common outer folds. This is a diagnostic using TER as teacher, not an
independently trained competing method.

\begin{table}[H]\centering\small
\caption{Affine re-expression on common outer folds (\%). The learned maps receive no additional observations.}
\label{tab:affine-projection}
\setlength{\tabcolsep}{4pt}
\begin{tabular*}{\linewidth}{@{\extracolsep{\fill}}lrrr@{}}
\toprule
Inputs and representation & PR-AUC & ROC-AUC & Macro-F1\\\midrule
Evidence (original) & 56.49 & 64.55 & 56.97\\
Evidence (affine map) & 59.43 & 67.70 & 56.90\\
Daily (original) & 54.89 & 64.48 & 56.68\\
Daily (affine map) & 59.21 & 67.77 & 57.35\\
Daily + evidence (original) & 56.49 & 65.41 & 57.02\\
Daily + evidence (affine map) & 61.36 & 70.37 & 58.40\\
\bottomrule\end{tabular*}
\end{table}

For daily plus evidence, the gains are 4.87/4.97/1.38 points, with 36/42 improved cells. For an original feature vector $u$ and affine export $z=Au+b$, a linear predictor of $z$ is still linear in $u$. The map changes the weighting of existing combinations and the regularization induced by the downstream penalty, without adding sample information. This demonstrates a useful re-expression of the same inputs, rather than uniquely identifying the cause of every remaining deficit.

The maps do not reproduce the teacher exactly. Fit/audit standardized teacher
MSE is .244/.491 for evidence, 1.358/1.780 for daily inputs, and .216/.462 for
their concatenation. Public-cohort MSE ranges are .454--1.662, 6.732--23.834,
and .887--1.602, respectively, under the same source-fitted target scales.
Neither teacher reconstruction error nor every task improves monotonically.
The remaining gap can involve nonlinear information use, omitted temporal
detail, or source-transfer error. The intervention establishes that changing
coordinates and their induced regularization can repair part of the deficit
without additional input information. Section~\ref{sec:observation-intervention}
separately varies which measurements are observed.

\begin{table}[H]\centering\footnotesize
\caption{Affine re-expression minus direct concatenation on all tasks (points).}
\label{tab:affine-task-delta}
\setlength{\tabcolsep}{4pt}
\begin{tabular*}{\linewidth}{@{\extracolsep{\fill}}lrrr@{}}
\toprule
Task & $\Delta$ PR & $\Delta$ ROC & $\Delta$ F1\\\midrule
CGMacros Diabetes & 4.83 & 3.38 & 1.05\\
CGMacros Insulin resistance & 0.44 & 0.99 & 2.11\\
CGMacros Hyperlipidemia & -15.76 & -16.07 & -16.42\\
CGMacros Obesity & 1.36 & 2.65 & 0.65\\
Shanghai Insulin resistance & 2.18 & 0.36 & -3.16\\
Shanghai Hyperlipidemia & 3.58 & 4.06 & -1.74\\
Shanghai Hypoglycemia & 3.30 & 2.73 & -3.68\\
Stanford Diabetes & 7.99 & 9.06 & 4.57\\
Stanford Beta-cell dysfunction & 5.71 & 10.24 & 6.15\\
Stanford Insulin resistance & 8.04 & 9.74 & 4.39\\
Hall Diabetes & 11.43 & 9.63 & 4.85\\
Hall Insulin resistance & 16.84 & 13.17 & 8.53\\
Hall Hyperlipidemia & 7.28 & 9.26 & 1.05\\
Hall Glucotype & 10.93 & 10.37 & 10.96\\
\bottomrule\end{tabular*}
\end{table}

\subsection{Separating observation changes from readout estimation}
\label{sec:observation-intervention}

The cross-group experiments vary who fits the readout. Here we instead hold
the fitted predictor fixed and vary which measurements it receives. Across all
2,708 public windows, we remove $\lfloor0.2n\rfloor$ of each window's $n$
observed points, either uniformly without replacement or as one circular run
through the ordered observed indices. Both modes remove the same count; four
prespecified masks per mode are seeded by window identity. Windows are verified
non-overlapping within recording identity and source. No labels enter masking.

We recompute daily representations, evidence, and TER from the retained readings
and masks, including interpolation, while preserving the original sampling
cadence. Thus deleted values cannot enter through a cached representation.
Unchanged-input extraction reproduces every archived feature array exactly.
This observation diagnostic uses the archived fixed-checkpoint $10\times5$
partitions and reports changes relative to its own unchanged inputs.
StandardScaler and LogisticRegression are
fitted once on unchanged training inputs and applied to the unchanged and
perturbed test inputs. Neither the encoders nor the affine map from
Section~\ref{sec:summary-projection} are updated. Every mask is scored separately;
scores are then averaged over masks and tasks, without prediction ensembling.

\begin{table}[H]\centering\small
\caption{\textbf{Observation sensitivity with a fixed downstream predictor.}
Metric changes are percentage points relative to unchanged input. $D$ is mean squared probability change from the
unchanged input; $V$ is prediction variance across the four masks, both
averaged over classes, test windows, folds, and tasks. Lower $D,V$ indicate
less prediction movement, not necessarily higher task accuracy.}
\begin{tabular}{llrrrrr}\toprule
Input & Export & $\Delta$ PR & $\Delta$ ROC & $\Delta$ F1 & $D$ & $V$\\\midrule
Scattered & Evidence & -1.19 & -1.66 & -11.09 & .2095 & .0084\\
 & Daily + evidence & -1.73 & -1.77 & -11.14 & .2245 & .0186\\
 & Affine daily + evidence & -1.30 & -1.21 & -9.81 & .0756 & .0031\\
 & TER & -4.00 & -2.41 & -14.96 & .1104 & .0039\\\midrule
Contiguous & Evidence & -1.99 & -2.28 & -11.52 & .2101 & .0190\\
 & Daily + evidence & -2.42 & -2.61 & -11.23 & .2235 & .0218\\
 & Affine daily + evidence & -1.95 & -1.62 & -2.47 & .0230 & .0065\\
 & TER & -2.57 & -1.66 & -4.94 & .0310 & .0054\\\bottomrule
\end{tabular}
\end{table}

TER and the affine representation move less than direct summaries under both
interventions, but ranking and thresholded decisions respond differently.
TER's PR-AUC/ROC-AUC losses are 4.00/2.41 points for scattered deletion and
2.57/1.66 for contiguous deletion; its Macro-F1 losses are 14.96 and 4.94 points.
The affine map has smaller losses in all three metrics for both modes, whereas
direct evidence has smaller PR-AUC losses despite lower absolute scores.
This separates two properties: stable estimation of a readout across people,
as trained by fit--transfer, and stability of that fixed readout under changed
observations. The latter is intervention- and metric-dependent. The clean-input
affine repair and paired cross-group scoring experiments establish the main
representation-learning explanation; this diagnostic characterizes observation
sensitivity rather than attributing those gains to universal denoising.

\subsection{Retained individual differences versus fitting-group sensitivity}
\label{sec:contrast-retention}

Lower prediction variance could arise simply by shrinking every output. We
test this alternative on the same fixed evaluation people and 24 fitting-group
draws used above. Average windows within each person, then center each evidence
coordinate across people. Let $t$ be the observed individual contrasts, $p_j$
their predictions under fitting draw $j$, and $m=\mathbb E_jp_j$. Define
\[
T=\mathbb E[t^2],\quad P=\mathbb E[m^2],\quad
C=\mathbb E[mt],\quad N=\mathbb E_j\mathbb E[(p_j-m)^2].
\]
Expectations weight people and coordinates equally within each cohort. Gain
$C/T$ measures the amplitude recovered along the observed contrasts; alignment
$C/\sqrt{TP}$ measures directional agreement. The ratio $NT/C^2$ expresses
fitting-group variation relative to the recovered target-aligned contrast and
is invariant to uniform output rescaling. We compute these quantities within
cohort and report equal-cohort means for the 272 longitudinal/event targets.

\begin{table}[ht]\centering\small
\caption{\textbf{Stability is not explained by uniform shrinkage.}
Individual-contrast diagnostics; the last column is contrast MSE normalized by
$T$. Both the stability gain and the fidelity cost are reported.}
\begin{tabular}{llrrrr}\toprule
Target & Objective & Gain & Alignment & $NT/C^2$ & Total error\\\midrule
Evidence & Joint & .62338 & .77745 & .26527 & .50116\\
Evidence & Fit--transfer & .47508 & .73181 & .12111 & .50021\\
Raw & Joint & .58767 & .74900 & .30085 & .55059\\
Raw & Fit--transfer & .40286 & .67306 & .19107 & .58422\\\bottomrule
\end{tabular}
\end{table}

For evidence targets, fitting variation relative to recovered contrast is
54.34\% lower with TER; the direction agrees in three of four cohorts.
Gain is also smaller, and alignment is not uniformly improved. In the three
paired scoring-group interventions of Section~\ref{sec:fit-transfer-intervention},
the same scale-invariant ratio falls by 47.95/49.36/35.86\%, with agreement in
11 of 12 seed--cohort pairs, while total contrast error rises in each seed.
Cross-group training therefore changes the fidelity--stability trade-off;
it does not simply recover every observed difference more accurately. The
observed evidence is not noise-free physiological ground truth. The next
diagnostic tests which feature differences contribute to phenotype predictions.

\subsection{Feature associations that disappear across people}
\label{sec:feature-transfer}

\paragraph{Question and fixed protocol.}
We test whether a predictor uses feature differences that distinguish classes
only in its fitting people. We evaluate seven representation conditions, including
the four shown in Figure~\ref{fig:feature-transfer}, on all 14 tasks and the main-table
person-grouped $10\times5$ outer folds, StandardScaler and logistic regression
($C=1$). The affine representation is the completed source-only projection
from daily plus evidence in Section~\ref{sec:summary-projection}, not an
additional selected model. Mean--max and BiLSTM retain their original trained encoders. TER and joint reconstruction use their saved inner-selected candidate in each outer fold; this audit never selects a model using the directional diagnostic.

Fit the scaler and a PCA basis on each downstream training fold only. The
smallest leading subspace explaining at least 90\% of training feature
variance is the dominant subspace; its orthogonal complement is the residual
subspace. This single rule was fixed before inspecting the directional results.
Neither phenotype labels nor held-out rows determine the basis. Low variance
is not presumed to mean noise. For each class, compute its mean feature vector
minus the mean of the other classes, separately on fitting and held-out people.
Within each subspace, cosine similarity compares these two contrast vectors.
Binary tasks use positive minus negative; multiclass tasks average classes
equally. All contrasts are defined on the evaluated folds.

\paragraph{A feature statistic and an intervention.}
For subspace projector $Q$, let $d_F,d_T$ be the fitting and held-out class
contrasts, and $w$ the fitted classifier coefficient (class versus other-class
coefficients for multiclass). We report
\[
a_Q=\frac{(Qd_F)^\top Qd_T}{\|Qd_F\|\,\|Qd_T\|},\qquad
m_F=w^\top Qd_F,\qquad m_T=w^\top Qd_T.
\]
The first is a feature statistic, not a classification score. The latter two
measure how those directions contribute to fitted and held-out class separation.
Dominant and residual margin contributions add exactly to the full margin.
We then remove only the residual contribution to logits, keeping the original
classifier coefficients, intercept and scaler fixed. This functional deletion
test is not a retrained PCA model or a new main-table entry. Original logits,
probabilities, metrics and fold fingerprints are verified before deletion.

\begin{table}[H]\centering\footnotesize
\caption{Residual class differences and the effect of removing their logit contribution. Changes are percentage points; positive means deletion helps.}
\label{tab:spectral-aligned}
\setlength{\tabcolsep}{4pt}
\begin{tabular*}{\linewidth}{@{\extracolsep{\fill}}lrrrrrr@{}}
\toprule
Representation & $a_Q$ & $m_F$ & $m_T$ & $\Delta$ PR & $\Delta$ ROC & $\Delta$ F1\\\midrule
Direct evidence & 0.023 & 2.754 & 0.239 & 1.128 & 1.572 & 0.017\\
Daily + evidence & -0.000 & 2.790 & 0.009 & 2.562 & 3.135 & 1.544\\
Joint reconstruction & -0.005 & 2.623 & 0.118 & 2.749 & 1.890 & 1.680\\
Mean--max pooling & 0.014 & 3.383 & 0.288 & 2.542 & 0.973 & 1.072\\
BiLSTM pooling & 0.010 & 2.923 & 0.184 & 1.486 & 1.854 & 0.750\\
Same-input affine map & 0.356 & 1.342 & 0.727 & -5.697 & -6.405 & -9.866\\
TER & 0.300 & 1.614 & 0.744 & -5.249 & -4.770 & -9.075\\
\bottomrule\end{tabular*}
\end{table}

\begin{table}[H]\centering\footnotesize
\caption{Joint reconstruction and TER on every task: class-contrast alignment and PR-AUC change after residual removal.}
\label{tab:spectral-tasks}
\setlength{\tabcolsep}{4pt}
\begin{tabular*}{\linewidth}{@{\extracolsep{\fill}}lrrrr@{}}
\toprule
Task & Joint alignment & TER alignment & Joint change & TER change\\\midrule
CGMacros Diabetes & 0.076 & 0.616 & 4.797 & -10.410\\
CGMacros Insulin resistance & 0.056 & 0.583 & -0.775 & -2.509\\
CGMacros Hyperlipidemia & 0.036 & 0.248 & -2.953 & -3.370\\
CGMacros Obesity & -0.006 & 0.383 & 3.924 & -11.088\\
Shanghai Insulin resistance & -0.025 & -0.348 & 0.079 & 4.069\\
Shanghai Hyperlipidemia & -0.100 & 0.042 & 8.570 & -0.037\\
Shanghai Hypoglycemia & 0.047 & -0.008 & 0.538 & -11.154\\
Stanford Diabetes & -0.028 & 0.593 & 5.130 & -8.632\\
Stanford Beta-cell dysfunction & -0.086 & 0.575 & 4.239 & -0.640\\
Stanford Insulin resistance & -0.104 & 0.588 & 8.369 & -6.010\\
Hall Diabetes & -0.016 & 0.202 & 3.137 & -1.999\\
Hall Insulin resistance & -0.118 & 0.127 & 9.439 & -9.404\\
Hall Hyperlipidemia & 0.176 & 0.167 & -8.469 & -3.191\\
Hall Glucotype & 0.017 & 0.434 & 2.455 & -9.108\\
\bottomrule\end{tabular*}
\end{table}

Smaller feature variations in the direct and joint-reconstruction representations show little cross-group class agreement. TER retains positive agreement on average, and removing these directions reduces PR-AUC in 13/14 tasks. The variance rule chooses different ranks and energies across representations; no labels or test rows fit the PCA basis. Held-out labels enter only the descriptive class contrasts and final metric computation. Reconstructed full logits are checked against the corresponding original outer-fold predictions before deletion.

\end{document}